\pdfoutput=1

\documentclass[10pt,twocolumn,letterpaper]{article}

\usepackage{preprint}

\usepackage{booktabs}
\usepackage{multirow}
\usepackage{tabularx}
\usepackage[table]{xcolor}
\usepackage{amsmath}
\usepackage{amssymb}
\usepackage{bm}
\usepackage{xspace}
\usepackage{subcaption}

\newcommand{\miniheadline}[1]{\vspace{1.2mm}\noindent\textbf{#1.}}

\providecommand{\epsSeg}{}\renewcommand{\epsSeg}{\mbox{$\epsilon$-Seg}\xspace}
\providecommand{\epsSegpp}{}\renewcommand{\epsSegpp}{\mbox{$\epsilon$-Seg++}\xspace}

\makeatletter
\providecommand{\miniheadline}{}
\renewcommand{\miniheadline}{\@ifstar\miniheadline@star\miniheadline@num}
\def\miniheadline@num#1{%
  \par\vspace{1.2mm}\noindent\refstepcounter{subsection}%
  \textbf{\thesubsection.~#1.}}
\def\miniheadline@star#1{%
  \par\vspace{1.2mm}\noindent\textbf{#1.}}
\makeatother

\definecolor{linkblue}{rgb}{0.21,0.49,0.74}
\usepackage[pagebackref,breaklinks,colorlinks,allcolors=linkblue]{hyperref}
\hypersetup{
  pdftitle    = {epsilon-Seg++: Ensembling Across Hierarchy and Training Time for Label-Efficient EM Segmentation},
  pdfauthor   = {Edoardo Giacomello, Sheida Rahnamai Kordasiabi, Mehdi Seifi, Florian Jug},
  pdfsubject  = {Sparsely supervised semantic segmentation of volume electron microscopy},
  pdfkeywords = {semantic segmentation, electron microscopy, sparse supervision, label efficiency, ensembling, hierarchical VAE},
}

\title{\epsSegpp: Ensembling Across Hierarchy and Training Time\\for Label-Efficient EM Segmentation}

\author{%
Edoardo Giacomello\textsuperscript{1}\thanks{Equal contribution; listed in alphabetical order.}\hspace{1.2em}%
Sheida Rahnamai Kordasiabi\textsuperscript{1,2}\footnotemark[1]\hspace{1.2em}%
Mehdi Seifi\textsuperscript{1}\hspace{1.2em}%
Florian Jug\textsuperscript{1}\\[0.4em]
\textsuperscript{1}Human Technopole\hspace{2em}\textsuperscript{2}Technische Universit\"at Dresden\\[0.2em]
{\tt\footnotesize \{edoardo.giacomello, sheida.rahnamai, mehdi.seifi, florian.jug\}@fht.org}%
}

\begin{document}
\maketitle

\begin{abstract}
Semantic segmentation of electron microscopy (EM) volumes is a central step in quantitative cell biology, but dense voxel-wise annotation is prohibitively expensive.
\epsSeg showed that a hierarchical variational autoencoder trained on sparsely annotated voxels can yield competitive segmentations, but it classifies from its topmost latent level only and operates exclusively on 2D patches.
We introduce \epsSegpp, which places a semantic expert at every level of the hierarchy and consolidates their predictions with a parameter-free product of experts.
We further enable 3D inputs through a leakage-free sub-volume splitting protocol, which allows 2D and 3D models to be compared under identical sparse supervision.
Because validation labels are just as scarce as training labels, we treat checkpoint selection as part of the method and average the top-ranked checkpoints, a second ensemble, this time across training time.
On the BetaSeg FIB-SEM dataset, the multi-level experts improve 2D mean Dice by about $0.1$ at $2{,}500$ training labels.
With up to $600$ training labels, the 2D model matches or exceeds 3D \epsSeg at about one eleventh of its prediction cost.
Above roughly a thousand labels, 3D inputs dominate, and at matched annotation cost 3D \epsSegpp reaches a mean Dice of $0.905$ versus $0.880$ for D3PM, while predicting between $5.4$ and $27$ times faster (with our checkpoint ensemble or a single model, respectively).
Checkpoint ensembling matches single-best selection on average and markedly improves the worst case when the validation set is small.
\end{abstract}

\section{Introduction}
\label{sec:intro}
\begin{figure}[t]
    \centering
    \includegraphics[width=\linewidth]{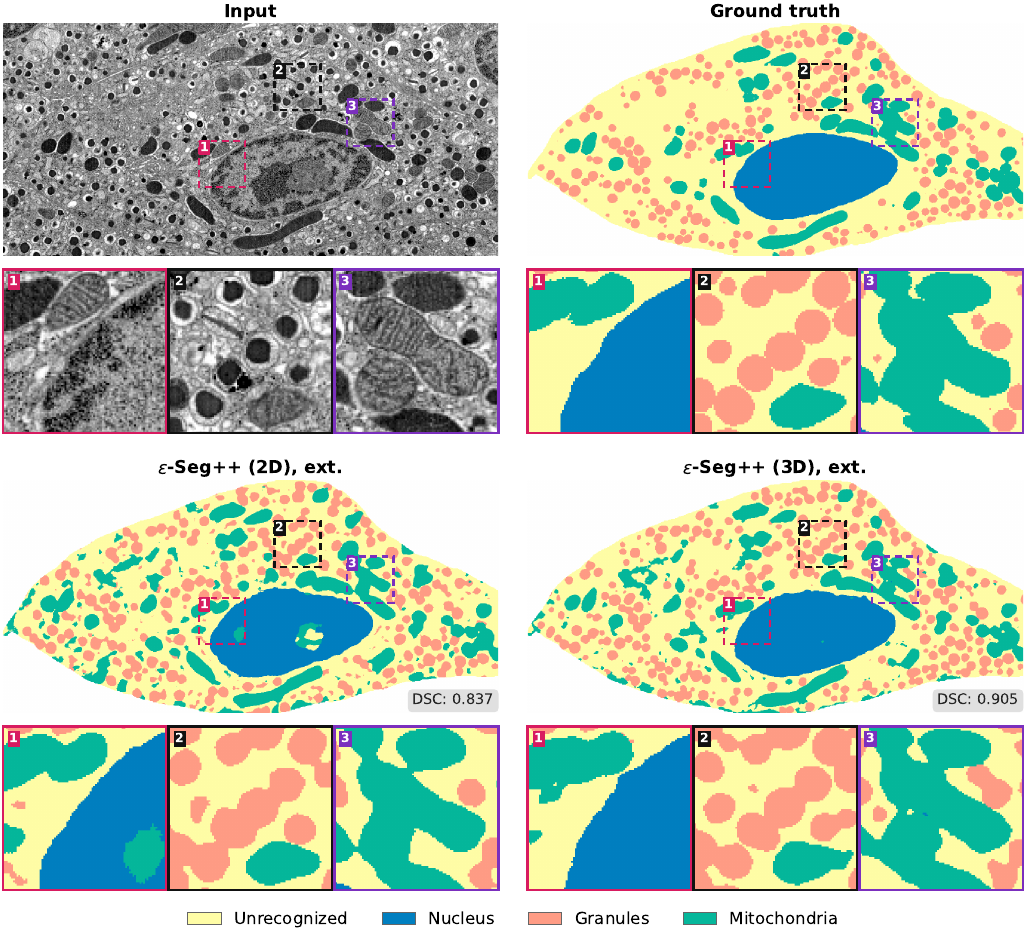}
    \caption{\textbf{Sparsely supervised segmentation of a FIB-SEM section.}
    Test section ($z=626$) of BetaSeg~\cite{betaseg}, ground truth, and predictions of 2D and 3D \epsSegpp trained with the same annotation as the baselines (Table~\ref{tab:main-epssegpp-results}).
    Mean Dice: $0.905$ (3D) and $0.837$ (2D), against $0.879$ and $0.822$ for \epsSeg; the 2D model predicts about eleven times faster.
    Insets show the nuclear boundary (1), granules (2), and mitochondria (3).}
    \label{fig:teaser}
\end{figure}

\begin{figure*}[t]
    \centering
    \captionsetup[subfigure]{justification=raggedright,singlelinecheck=false}
    \begin{subfigure}{0.980\textwidth}
        \caption{}\label{fig:arch_model}
        \includegraphics[width=\linewidth]{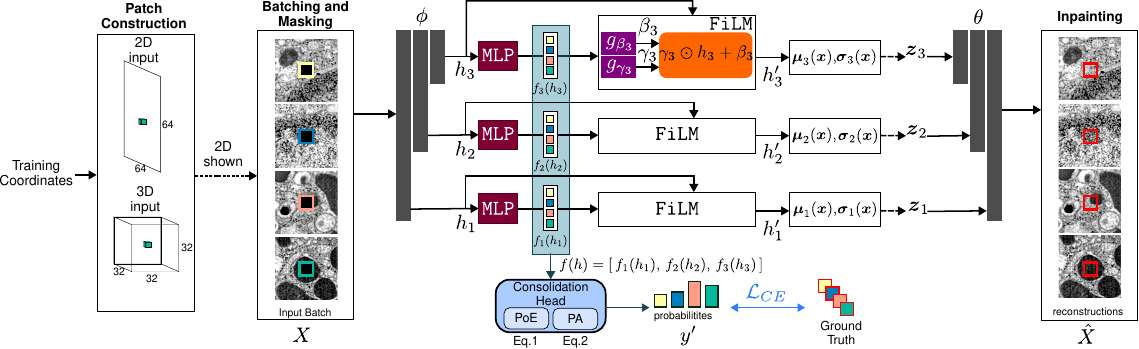}
    \end{subfigure}

    \vspace{2mm}
    \begin{subfigure}{0.559\textwidth}
        \caption{}\label{fig:arch_data}
        \includegraphics[width=\linewidth]{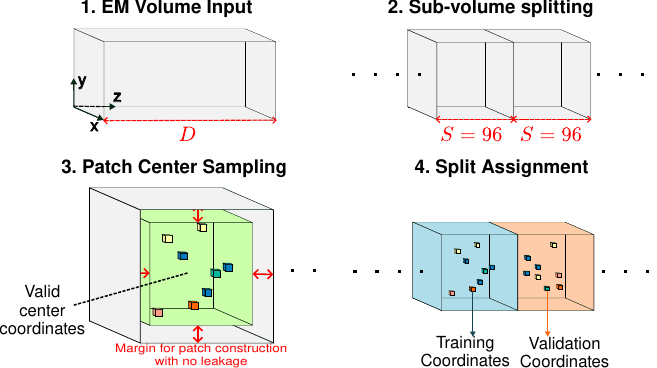}
    \end{subfigure}\hfill
    \begin{subfigure}{0.375\textwidth}
        \caption{}\label{fig:arch_selection}
        \includegraphics[width=\linewidth]{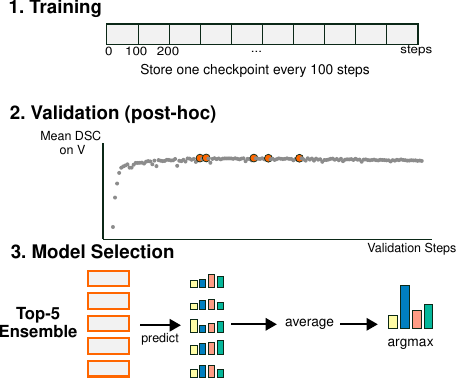}
    \end{subfigure}
\caption{\textbf{Overview of \epsSegpp.}
\textbf{(a) Architecture.} A 2D or 3D patch around the annotated voxel enters a three-level HVAE
with its central region zero-masked for reconstruction. Each level's expert predicts class logits
$f_\ell(h_\ell)$ and, through FiLM, conditions that level's Gaussian posterior. The consolidation
head merges the $L=3$ experts into one prediction $y'$, by Product of Experts
(Eq.~\eqref{eq:poe}, default) or posterior averaging (Eq.~\eqref{eq:smv}); the cross-entropy
$\mathcal{L}_{CE}$ is applied once, to $y'$, at annotated coordinates only.
\textbf{(b) Leakage-free data preparation.} The volumes used for training and validation are cut into sub-volumes of depth
$S$, assigned whole to training or validation; patch centres are sampled only where no context up to
$\mathbf{P}_{\max}$ crosses a boundary.
\textbf{(c) Training and selection.} A checkpoint is stored every $100$ steps, with no validation
in the loop; all are scored \textit{post hoc} on the validation set $V$, and the posteriors of the
best $N=5$ are averaged (Eq.~\eqref{eq:ensemble}) before $\arg\max$.}
\label{fig:epssegpparch}
\end{figure*}

Volume electron microscopy (vEM) now routinely images whole cells and tissues at nanometre resolution~\cite{peddie2022vem}.
Turning such volumes into quantitative biology, \eg, measuring how organelles are distributed, how they contact each other, or how they change in disease, requires a semantic segmentation of every voxel~\cite{heinrich2021cosem,betaseg}.
Supervised networks can deliver this with high accuracy~\cite{unet,heinrich2021cosem}, but only after experts have densely annotated large amounts of training data, an effort that has to be repeated for every new sample type, imaging setup, or set of classes of interest.
In practice, the annotation a biologist can afford is sparse: individual voxels scattered over the structures of interest rather than densely labelled regions.

Existing methods put such scarce labels to use in different ways.
Self-supervised approaches learn voxel representations without labels and assign classes afterwards, \eg, by clustering~\cite{han,maester}, but the groups they recover need not coincide with the classes a biologist is interested in.
Sparsely supervised networks propagate a few labels to unlabelled pixels~\cite{agmm,d3pm}, but they operate on patches and require at least one label for every class present in each training patch.
\epsSeg~\cite{epsseg} instead makes the individual voxel the unit of annotation.
A hierarchical variational autoencoder is trained jointly to inpaint a masked centre region of the patches built around the annotated voxels and to classify the centre voxel of each patch, so that every annotated voxel can sit anywhere in the volume and is used directly.

Three things limit how far this regime can be pushed.
First, \epsSeg reads its prediction from the topmost level of its hierarchy only, leaving unused the semantic information carried by the lower levels.
Second, \epsSeg operates on 2D patches of what are 3D volumes.
Moving to volumetric inputs is simple architecturally, but when sparse training and validation labels are drawn from the same volumes, the context of a training patch can overlap that of a validation patch, and a naive split leaks information between the two.
Third, and common to the whole field, model selection needs labels too.
Choosing a checkpoint, a stopping point, or a hyperparameter requires a validation set, which in the sparse regime is drawn from the same annotation budget, yet validation labels are rarely counted as annotation cost and can outnumber the training labels being reported~\cite{oliver2018realistic}.

We address these points with \epsSegpp, whose central idea is to ensemble twice.
Across the hierarchy, we attach a semantic expert to every latent level, let each expert condition the posterior of its own level as in \epsSeg, and fuse the experts with a parameter-free product of experts~\cite{hinton2002poe}.
The experts receive no loss of their own and are trained only through their joint prediction (Section~\ref{sec:multilevel}).
Across training time, we rank the stored checkpoints on the validation set and average the predictions of the five best, which consumes exactly the annotations needed to pick the single best one (Section~\ref{sec:selection}).
To make volumetric inputs usable, we split the data into sub-volumes and draw a single pool of annotations that serves 2D and 3D models alike, so that both are compared under identical supervision (Section~\ref{sec:dataset_preprocessing}).
Figure~\ref{fig:teaser} shows the resulting segmentations.

On the BetaSeg FIB-SEM dataset~\cite{betaseg}, these choices change what a given annotation budget buys.
With the validation set held fixed, the multi-level experts shift the 2D label-efficiency curve along the annotation axis: \epsSegpp reaches with $2{,}500$ training labels the Dice score that \epsSeg needs about $9{,}000$ for.

Up to $600$ training labels, the 2D model matches or exceeds 3D \epsSeg while predicting about eleven times faster.
Above roughly a thousand labels the volumetric input takes over.
At matched annotation cost, 3D \epsSegpp reaches a mean Dice of $0.905$, against $0.880$ for D3PM~\cite{d3pm} and $0.769$ for AGMM-SASS~\cite{agmm}.
Most of this comes from the volumetric input, which lifts \epsSeg from $0.822$ to $0.879$, on par with D3PM, and averaged over three seeds the experts add to it at most budgets (Section~\ref{sec:exp_label_efficiency}).
With about two and a half times fewer annotations, our 3D models already match D3PM to within $0.006$ mean Dice.
Finally, how validation labels are spent matters as much as how many there are.
For a training budget of $20{,}000$ labels, a validation set of $5{,}000$ labels combined with checkpoint ensembling selects a better model ($0.835$) than the single best checkpoint on a pool of $98{,}000$ ($0.817$).
Across all budgets, the ensemble changes the average result only moderately, if at all, but consistently raises the worst case, which is what a user with one small validation set wants.

Code, models and all coordinate sets used in this work will be made publicly available.

\section{Related Work}
\label{sec:related}

\miniheadline*{Sparse and weak supervision for segmentation}
Fully supervised networks segment EM organelles accurately~\cite{unet,heinrich2021cosem}, but at the cost of dense voxel-wise annotation.
Weakly supervised methods reduce this cost with scribbles~\cite{scribblesup}, points~\cite{bearman2016point}, or a few annotated slices per volume~\cite{cicek2016unet3d}.
In microscopy, interactive tools train lightweight classifiers on hand-crafted features from user scribbles~\cite{ilastik,labkit}, and promptable foundation models speed up the annotation of individual instances~\cite{microsam}.
Self-supervised voxel representations, learned without labels and grouped afterwards, rediscover several subcellular structures in vEM~\cite{han,maester}, and large unlabelled EM corpora support pre-training~\cite{conrad2021cem500k}.
Closest to our setting are two sparsely supervised methods.
AGMM-SASS~\cite{agmm} builds class-wise Gaussian mixtures around the features of labelled pixels and uses them to supervise nearby unlabelled pixels.
D3PM~\cite{d3pm} casts segmentation from a single labelled pixel as conditional discrete diffusion and aggregates sampled point clouds into a mask.
Both annotate per patch and require at least one label for every class present in each training patch, whereas \epsSeg~\cite{epsseg} and \epsSegpp annotate individual voxels.
Because the two units do not coincide, we compare against them at matched annotation cost (Section~\ref{sec:exp_baselines}).

\miniheadline*{Hierarchical latent variable models}
\epsSeg builds on hierarchical variational autoencoders~\cite{lvae,nvae,vdvae} in the form used for microscopy image restoration~\cite{hdn}, trained to inpaint a masked central region, an objective related to masked image modelling~\cite{mae,maester}.
Sparse labels enter through a top-level Gaussian-mixture prior with one block of latent channels per class, a contrastive term on the posterior means, and a cross-entropy on the centre voxel.
We keep all of these and change only where and how the semantic prediction is read out.

\miniheadline*{Multi-level prediction and fusion of experts}
Reading pixel-wise predictions from several levels of a network is well established: hypercolumns~\cite{hypercolumns} concatenate features across depth, and deeply supervised networks~\cite{dsn} attach a loss to intermediate layers.
Our experts differ in two respects.
Each expert conditions the posterior of its own level through FiLM~\cite{film}, so the semantic prediction shapes latent inference instead of only reading from it.
And no expert receives a loss of its own, so the experts are trained only through their fused prediction.
For the fusion we compare a product of experts~\cite{hinton2002poe}, which sums logits and lets confident experts dominate, with an average of the expert distributions~\cite{jacobs1991moe}; the product is never substantially worse in our ablations, and we use it as our default.

\miniheadline*{Model selection under scarce validation data}
Averaging the predictions of several models reliably improves accuracy and calibration~\cite{deepens}, and snapshot~\cite{snapshot} or checkpoint ensembles~\cite{checkpointens}, as well as weight averaging~\cite{swa}, obtain such ensembles from a single training run.
Less attention has been paid to what model selection costs.
In semi-supervised learning, Oliver~\etal~\cite{oliver2018realistic} showed that validation sets are often larger than the labelled training sets they are used to tune, and that realistically small validation sets make model selection unreliable, in line with the large error bars of small evaluation sets in general~\cite{varoquaux2018cv}.
Leakage between data splits is a further, widespread source of over-optimistic results~\cite{kapoor2023leakage}, and it becomes acute for patch-based models whose input context can cross split boundaries.
We address both together: our splits are leakage-free by construction, we count validation labels as part of the annotation budget, and we ask how a fixed budget is best divided between training and validation, and how the checkpoint used for prediction should then be chosen.

\section{Methods}
\label{sec:methods}

\miniheadline{Sparse Segmentation using Hierarchical Masked Autoencoders}
As in \epsSeg~\cite{epsseg}, annotations are available only at a sparse set of individual voxels, which we also refer to as \emph{patch centres}, since every training and validation patch is centred on one; the model receives the patch around such a voxel and predicts the class of its centre pixel or voxel, and at inference it is applied at every valid location.
We retain the main principles of \epsSeg, \ie, a hierarchical variational autoencoder (HVAE) backbone, centre-mask inpainting, a class-structured Gaussian-mixture prior at the topmost latent level, contrastive regularisation of the latent representations and direct semantic prediction, and refer to~\cite{epsseg} for their complete formulation.
All of these terms, the inpainting objective included, are computed on the patches around the training annotations alone, \ie, no pseudo-label is assigned to any unlabelled location (Supp.\ Sec.~\ref{sec:training_details}).
\epsSegpp goes beyond \epsSeg with two extensions that can be adopted independently (Section~\ref{sec:exp_architecture}): semantic experts at every level of the hierarchy (Section~\ref{sec:multilevel}) and volumetric inputs (Section~\ref{sec:volumetric}).
To compare 2D and 3D models under identical supervision, we construct spatially disjoint data splits and reuse the same patch centres for both (Figure~\ref{fig:epssegpparch}).

\miniheadline{Volumetric Inputs}
\label{sec:volumetric}
For 3D inputs, the input patch becomes a cube and every spatial operator of the backbone is replaced by its volumetric counterpart; all other architectural choices, loss terms and weights are identical in the two configurations.
The 2D models take a $64^2$ patch and the 3D models a $32^3$ one, so the 3D model sees eight times as many voxels but a quarter of the in-plane area, trading in-plane extent for context across $32$ consecutive sections (Supp.\ Sec.~\ref{sec:volumetric_details}).

\miniheadline{Leakage-free Data Preparation}
\label{sec:dataset_preprocessing}
In \epsSeg, training and validation patch centres were assigned per 2D section, which is safe for 2D inputs but not for 3D ones, whose context extends into neighbouring sections and can overlap the context of validation patch centres.
We therefore cut each volume used for training and validation into sub-volumes of depth $S$ and assign whole sub-volumes to training or validation by stratified $K$-fold over the original volumes, so that each volume contributes to both splits.
Patch centres are sampled only at locations where no input context up to a maximum patch size $\mathbf{P}_{\max}$ crosses a sub-volume boundary, and setting $\mathbf{P}_{\max}$ to the largest patch of either configuration lets a single pool of patch centres serve 2D and 3D models alike.
Normalisation statistics are computed on the training sub-volumes only.
Supp.\ Sec.~\ref{sec:suppl_preprocessing} gives the full procedure.

\miniheadline{Multi-level semantic experts}
\label{sec:multilevel}
In \epsSeg \cite{epsseg}, the semantic prediction is produced by a single classifier placed at
the \emph{top} level of the hierarchy. In particular, the features computed at that level are
passed through a small classification head that outputs one logit vector $f(h_L)\in\mathbb{R}^C$,
with $C$ being the number of semantic classes, and the class of the centre voxel is produced directly
from those logits. The same logits are then re-used to condition the latent inference of that
level: a FiLM layer~\cite{film}, modulates the level features before these are turned into the parameters of the latent posterior.
The remaining levels $\ell < L$ carry no classifier, and their posteriors are inferred
from their own features alone.

\epsSegpp applies the same construction to every level of the hierarchy. In particular, each of
the $L$ stochastic blocks carries its own classifier, which conditions the latent inference of its
own level exactly as in \epsSeg, and at the same time produces a prediction for the centre voxel.
This means that a single forward pass now yields $L$ class predictions instead of one,
\[
  f(h) = \big(f_1(h_1), f_2(h_2), \ldots, f_L(h_L)\big),
\]
where $h_\ell$ are the features available at level $\ell$, and $L=3$ in all our experiments.
We expect these predictions to be complementary, since the levels read features at different resolutions and degrees of abstraction.
For this reason we treat them as a set of \emph{experts} on the same voxel, and we need a rule to
consolidate them into the single label that \epsSegpp finally outputs.

We consider two different fusion strategies, the product and the sum rule of classifier combination~\cite{kittler1998combining}: Product of Experts (PoE)~\cite{hinton2002poe} and posterior averaging (PA)~\cite{jacobs1991moe}.
PoE sums expert logits before softmax,
\begin{equation}
p^{PoE}(c\mid x)=
\operatorname{softmax}_c\left(
\sum_{\ell=1}^{L}f_\ell(h_\ell)
\right),
\label{eq:poe}
\end{equation}
whereas PA averages the expert distributions
\begin{equation}
p^{PA}(c\mid x)=
\frac{1}{L}\sum_{\ell=1}^{L}
\operatorname{softmax}_c(f_\ell(h_\ell)).
\label{eq:smv}
\end{equation}

\miniheadline{Objective}
In \epsSegpp, supervision is applied only once, to the aggregated distribution of Eq.~\eqref{eq:poe} or Eq.~\eqref{eq:smv}. We add no classification loss to the individual levels, so the per-level classifiers are trained only through the prediction they produce together. The objective is
\begin{equation}
  \mathcal{L} = \alpha\,\mathcal{L}_{\mathrm{inp}}
              + \beta \sum_{\ell=1}^{L}\mathrm{KL}_\ell
              + \gamma\,\mathcal{L}_{\mathrm{CL}}
              + \mathcal{L}_{\mathrm{CE}},
\end{equation}
where $\mathcal{L}_{\mathrm{inp}}$ is the negative log-likelihood of the masked centre region,
$\mathcal{L}_{\mathrm{CL}}$ is the contrastive term of \epsSeg~\cite{epsseg}, computed from the
posterior means of all $L$ levels, and $\mathcal{L}_{\mathrm{CE}}$ is the cross-entropy on the
aggregated prediction. In all our experiments we set $\alpha = 1$, $\beta = 10^{-2}$ and $\gamma = 10^{-1}$.

\miniheadline{Latent Feature Cropping}
Since each expert reads the full feature map of its level, it can use features far from the voxel it classifies.
We therefore also test restricting each expert to a centred crop of $50\%$ or $25\%$ of its feature map, while the backbone and the latent inference still see the whole map (Section~\ref{sec:exp_architecture}, Supp.\ Tab.~\ref{tab:abl-feature-size}).

\miniheadline{Checkpoint Selection}
\label{sec:selection}
Since validation labels are scarce, the choice of checkpoint matters, and we treat it as a separate step that operates \textit{post hoc} on the checkpoints a training run has stored.
Let $\theta_1,\ldots,\theta_M$ be the stored checkpoints, and let $s(\theta; V)$ be the Dice score
obtained by $\theta$ on a validation set $V$. We order the checkpoints so that
$s(\theta_{(1)}; V)\ge\ldots\ge s(\theta_{(M)}; V)$, and we consolidate the $N$ best of them by
averaging their aggregated distributions,
\begin{equation}
p^{\mathrm{ens}}(c\mid x)=\frac{1}{N}\sum_{m=1}^{N} p\big(c\mid x,\theta_{(m)}\big),
\label{eq:ensemble}
\end{equation}
where $p(c\mid x,\theta)$ is the distribution that the checkpoint $\theta$ produces through
Eq.~\eqref{eq:poe} or Eq.~\eqref{eq:smv}, and the final label is
$\arg\max_c p^{\mathrm{ens}}(c\mid x)$. We use $N=5$ in all our experiments.
For our final models and the validation-budget study, training never sees the validation set; the fixed learning-rate schedule was fitted once to earlier runs (Supp.\ Sec.~\ref{sec:training_details}).
Eq.~\eqref{eq:ensemble} is parameter-free, requires no retraining, and consumes exactly the same annotations as selecting the single best checkpoint on $V$, since both need the same ranking.
\epsSegpp thus averages class posteriors twice: across the levels of the hierarchy and across training time.

\section{Experimental Setup and Results}
\begin{table*}[t]
\centering
\small
\setlength{\tabcolsep}{4pt}
\renewcommand{\arraystretch}{0.9}
\begin{tabular}{l!{\color{gray!50}\vrule}ccc!{\color{gray!50}\vrule}ccccc!{\color{gray!50}\vrule}c}
\hline
Model & $\lvert T\rvert$ & $\lvert V\rvert$ & $\lvert T\rvert+\lvert V\rvert$ & U & N & G & M & Avg. Dice & Pred. time \\
\hline
\multicolumn{10}{l}{\textit{Baselines: one labelled voxel per class present in each of $10K$ training and $6K$ validation patches}} \\
AGMM-SASS~\cite{agmm} & $25{,}823$ & $16{,}069$ & $41{,}892$ & $0.831$ & $0.724$ & $0.741$ & $0.779$ & $0.769$ & $1.0$\,s \\
D3PM~\cite{d3pm}      & $25{,}823$ & $16{,}069$ & $41{,}892$ & $0.897$ & $0.957$ & $0.809$ & $\mathbf{0.857}$ & $0.880$ & $14.7$\,h \\
\hline
\multicolumn{10}{l}{\textit{Ours, same annotation as the baselines}} \\
2D \epsSeg & $25{,}822$ & $16{,}069$ & $41{,}891$ & $0.834$ & $0.978$ & $0.716$ & $0.762$ & $0.822$ & $14.5$\,min \\
3D \epsSeg & $25{,}822$ & $16{,}069$ & $41{,}891$ & $0.889$ & $0.976$ & $0.880$ & $0.772$ & $0.879$ & $2.7$\,h \\
\rowcolor{gray!15}
2D \epsSegpp & $25{,}822$ & $16{,}069$ & $41{,}891$ & $0.863$ & $0.947$ & $0.787$ & $0.751$ & $0.837$ & $14.5$\,min \\
\rowcolor{gray!15}
3D \epsSegpp & $25{,}822$ & $16{,}069$ & $41{,}891$ & $\mathbf{0.916}$ & $\mathbf{0.982}$ & $\mathbf{0.901}$ & $0.822$ & $\mathbf{0.905}$ & $2.7$\,h \\
\hline
\multicolumn{10}{l}{\textit{Ours, same training annotation, validated on patch centres only}} \\
2D \epsSeg & $25{,}822$ & $6{,}124$ & $31{,}946$ & $0.833$ & $0.975$ & $0.724$ & $0.740$ & $0.818$ & $14.5$\,min \\
3D \epsSeg & $25{,}822$ & $6{,}124$ & $31{,}946$ & $0.889$ & $0.976$ & $0.880$ & $0.772$ & $0.879$ & $2.7$\,h \\
\rowcolor{gray!15}
2D \epsSegpp & $25{,}822$ & $6{,}124$ & $31{,}946$ & $0.860$ & $0.945$ & $0.774$ & $0.750$ & $0.832$ & $14.5$\,min \\
\rowcolor{gray!15}
3D \epsSegpp & $25{,}822$ & $6{,}124$ & $31{,}946$ & $0.913$ & $0.983$ & $0.895$ & $0.821$ & $0.903$ & $2.7$\,h \\
\hline
\multicolumn{10}{l}{\textit{Ours, patch centres only (same patch locations as the baselines)}} \\
2D \epsSeg & $10{,}000$ & $6{,}124$ & $16{,}124$ & $0.782$ & $0.936$ & $0.657$ & $0.737$ & $0.778$ & $14.5$\,min \\
3D \epsSeg & $10{,}000$ & $6{,}124$ & $16{,}124$ & $0.894$ & $0.967$ & $0.861$ & $0.808$ & $0.883$ & $2.7$\,h \\
\rowcolor{gray!15}
2D \epsSegpp & $10{,}000$ & $6{,}124$ & $16{,}124$ & $0.841$ & $0.944$ & $0.758$ & $0.752$ & $0.824$ & $14.5$\,min \\
\rowcolor{gray!15}
3D \epsSegpp & $10{,}000$ & $6{,}124$ & $16{,}124$ & $0.891$ & $0.954$ & $0.884$ & $0.767$ & $0.874$ & $2.7$\,h \\
\hline
\end{tabular}
\caption{\textbf{Comparison with the baselines at controlled annotation cost.}
         Mean and per-class Dice on the common region of the test section; $\lvert T\rvert$ and $\lvert V\rvert$ count annotated training and validation voxels (Supp.\ Sec.~\ref{sec:baselines_details}).
         With the same annotation as the baselines, which need a label for every class present in each patch, 3D \epsSegpp is best on three of the four classes and on average (bold: best of this matched comparison); the last two blocks reduce our annotation (Section~\ref{sec:exp_baselines}).
         Baselines are reported at their last checkpoint, our models (grey: \epsSegpp) with the top-$5$ checkpoint ensemble (other selection rules in Supp.\ Tab.~\ref{tab:strategy}).
         Each row is a single run; Figure~\ref{fig:label_efficiency} and Supp.\ Tab.~\ref{tab:abl-architecture} compare our models over several seeds.
         Prediction time is for one section and includes the five ensemble passes for our models.
         The two 3D \epsSeg rows that share an annotation budget are identical because both validation sets select the same five checkpoints, and the ensemble averages them regardless of order.}
\label{tab:main-epssegpp-results}
\end{table*}

\miniheadline{Dataset, Splitting and Pre-Processing}
We use the BetaSeg FIB-SEM dataset~\cite{betaseg} and, following prior work~\cite{epsseg}, the classes nucleus (N), granules (G) and mitochondria (M), with the remaining four merged into an Unrecognised class (U); voxels outside the cell are excluded from every loss and metric.
Three volumes (\textit{high\_c1--c3}) provide training and validation data, \textit{high\_c4} the test set.

\miniheadline{Annotation Pool}
The protocol of Section~\ref{sec:dataset_preprocessing}, with $S=96$, $K=3$ and $\mathbf{P}_{\max}=[32,64,64]$, yields a pool of $\lvert T\rvert = 160{,}000$ training and $\lvert V\rvert = 98{,}000$ validation annotations, from which all budgets below are drawn as nested, class-balanced subsets (Supp.\ Sec.~\ref{sec:dataset_details}).

\miniheadline{Evaluation}
We report per-class Dice scores and their unweighted mean on section $z=626$ of the test volume, so we stay comparable to methods previously tested on the same section~\cite{epsseg}.
Since 2D and 3D models predict slightly different regions, we score every experiment on their \emph{common region} of $311{,}821$ voxels (Supp.\ Sec.~\ref{sec:dataset_details}).

\miniheadline{Comparison against Baselines}
\label{sec:exp_baselines}
We fix a budget of $\lvert T\rvert = 10{,}000$ training and $\lvert V\rvert = 6{,}124$ validation patch centres.
Both are nested, class-balanced subsets of the pool drawn with a fixed seed, $2{,}500$ training centres per class, and the validation size keeps the pool's validation-to-training ratio.
AGMM-SASS and D3PM annotate per patch and need at least one label for every class present in a patch, so we add one labelled voxel per class present in each of these patches, which yields the $25{,}822$ training and $16{,}069$ validation voxels on which we train these baselines.
We compare our models in three configurations with decreasing annotation: with the full expanded annotation (same annotation as the baselines), with the expanded training set but validated on the original patch centres, and with the original patch centres alone (same patch locations as the baselines).
The comparison thus matches annotation, the quantity that has to be paid for, rather than the image data each method sees (Supp.\ Sec.~\ref{sec:baselines_details}).

\begin{figure*}[t]
    \centering
    \includegraphics[width=\textwidth]{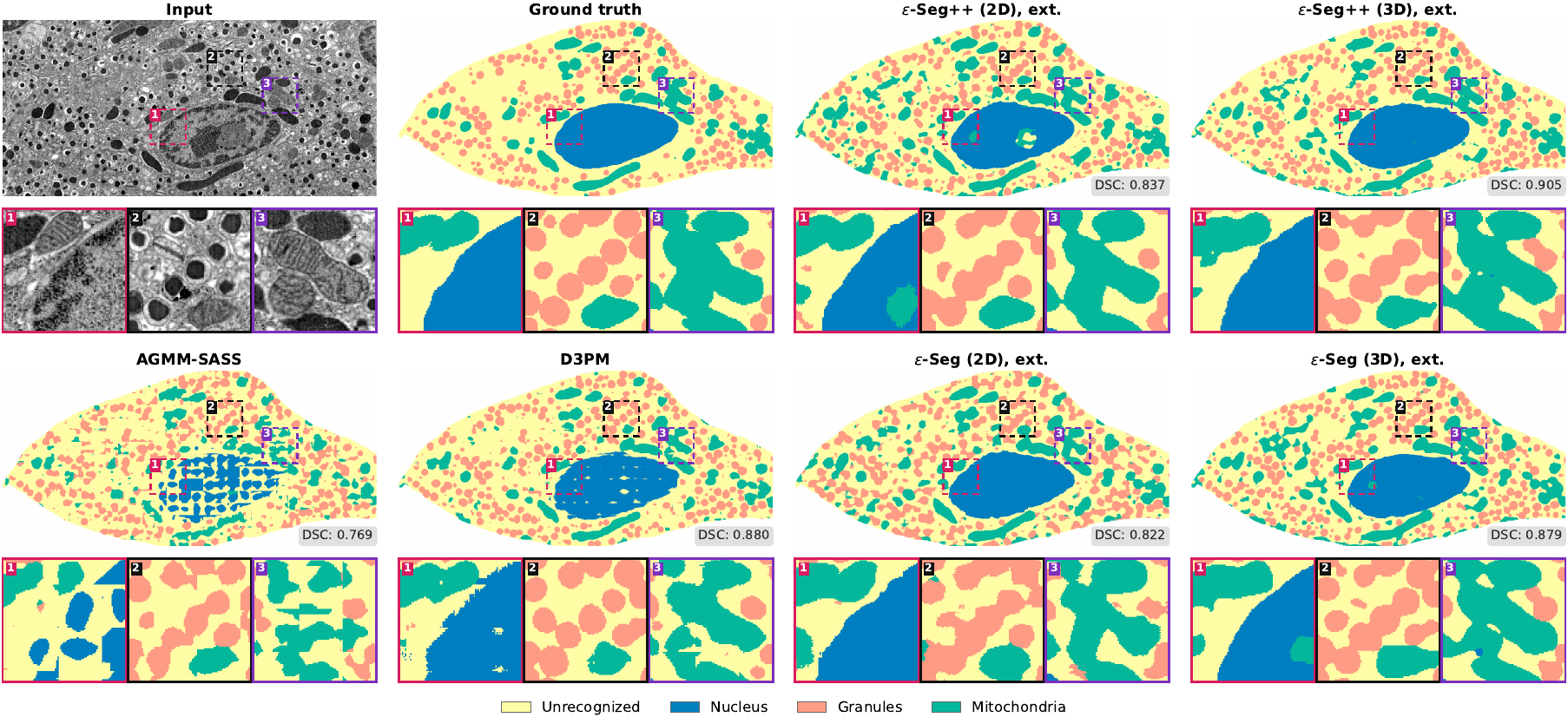}
    \caption{\textbf{Qualitative results of the baselines and our models at matched annotation.}
    All models are trained with the same annotation as the baselines ($\lvert T\rvert \approx 26$K and $\lvert V\rvert \approx 16$K annotated voxels; first block of ours in Table~\ref{tab:main-epssegpp-results}).
    Our models show the top-$5$ checkpoint ensemble selected on validation Dice, AGMM-SASS~\cite{agmm} and D3PM~\cite{d3pm} their last checkpoint.
    Insets show the nuclear boundary (1), granules (2), and mitochondria (3).
    Supp.\ Fig.~\ref{fig:supp_qualitative} shows all configurations of Table~\ref{tab:main-epssegpp-results}.}
    \label{fig:seg_models}
\end{figure*}

At matched annotation cost (about $42$K voxels), 3D \epsSegpp reaches $0.905$ mean Dice, against $0.880$ for D3PM and $0.769$ for AGMM-SASS (Table~\ref{tab:main-epssegpp-results}); the gain is largest on granules ($0.901$ against $0.809$), while D3PM keeps an edge on mitochondria ($0.857$ against $0.822$).
Most of this comes from the volumetric input (\epsSeg: $0.822 \rightarrow 0.879$, on par with D3PM); the experts add $0.026$ in 3D and $0.015$ in 2D (all our models use the same checkpoint ensemble).
Validating on the original patch centres only costs 3D \epsSegpp $0.002$ mean Dice and saves $24\%$ of the total annotation (Section~\ref{sec:exp_selection}).
With the original patch centres alone, \ie, about two and a half times less annotation, 3D \epsSegpp and 3D \epsSeg reach $0.874$ and $0.883$, within $0.006$ mean Dice of D3PM or above it.
This single-run gap is within the seed-to-seed spread of our 3D models (Supp.\ Tab.~\ref{tab:abl-architecture}).
D3PM needs $14.7$\,h to predict one section, against $2.7$\,h and $14.5$\,min for our 3D and 2D models including the ensemble, whereas AGMM-SASS segments a section densely in about a second (Supp.\ Tab.~\ref{tab:runtime}); Figure~\ref{fig:seg_models} shows qualitative results.

\miniheadline{\epsSegpp Architecture Selection}
\label{sec:exp_architecture}
We select the architecture at a low training budget of $\lvert T\rvert = 2{,}500$ annotations, using the full validation pool of $\lvert V\rvert = 98{,}000$ to minimise the influence of model selection (details in Supp.\ Sec.~\ref{sec:ablation_details}).
In 2D, placing an expert at every level improves over \epsSeg for both fusion rules, under all three selection metrics and for all five seeds, by up to $0.111$ mean Dice ($0.653 \rightarrow 0.764$, Supp.\ Tab.~\ref{tab:abl-architecture}).
In 3D, where \epsSeg is already strong at this budget ($0.829$), the gain depends on the selection metric.
The Product of Experts is at least as good as posterior averaging in $11$ of $12$ comparisons (Supp.\ Tab.~\ref{tab:abl-aggregation}), and we use it as our default.
The best feature-crop size depends on the budget (Supp.\ Tab.~\ref{tab:budget-sweep}); based on the label-efficiency experiment, we use a $50\%$ crop in 2D and the full feature map in 3D.
All three choices also follow from validation Dice alone: the experts improve 2D \epsSeg for all seeds and metrics, the Product of Experts is at least as good as posterior averaging in all $12$ comparisons, and in 2D the $50\%$ crop is ahead of the full feature map at all twelve label-efficiency budgets (Supp.\ Sec.~\ref{sec:ablation_details}).

\miniheadline{Label Efficiency}
\label{sec:exp_label_efficiency}
\begin{figure*}[t]
    \centering
    \includegraphics[width=\textwidth]{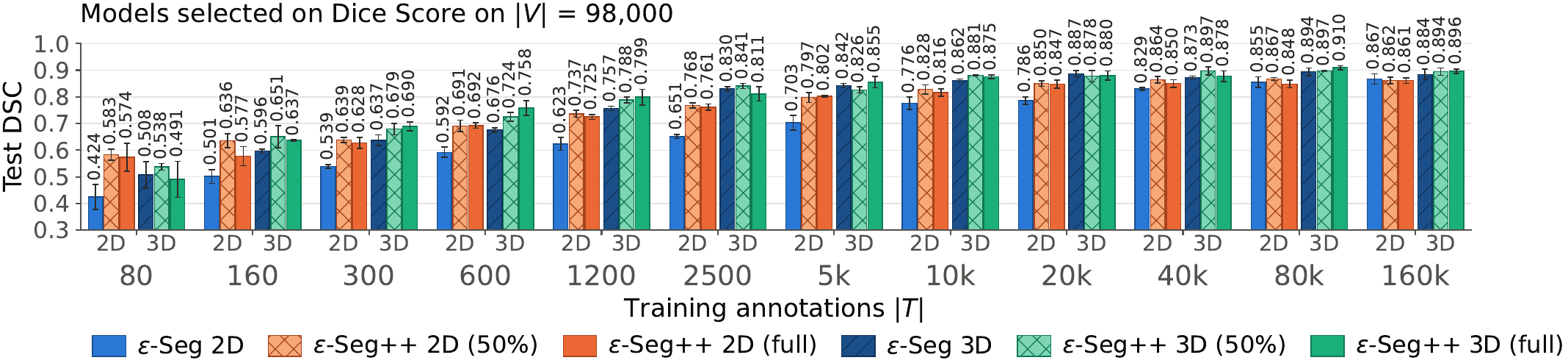}
    \caption{\textbf{Label efficiency of \epsSeg and \epsSegpp.}
    Mean Dice on the test section ($z=626$) for nested training sets of $80$ to $160{,}000$ class-balanced annotations, with all models selected by Dice on the same validation pool of $\lvert V\rvert = 98{,}000$ annotations.
    We compare \epsSeg and \epsSegpp in 2D and 3D, with \epsSegpp reading either the full feature map or a $50\%$ crop of it; each model is trained with three seeds.
    Supp.\ Fig.~\ref{fig:supp_label_efficiency} shows the same experiment under other selection rules.
    }
    \label{fig:label_efficiency}
\end{figure*}

To measure how efficiently each architecture uses a training budget, we train on $12$ nested training sets of $80$ to $160{,}000$ annotations, keeping the full validation pool of $\lvert V\rvert = 98{,}000$ fixed to minimise the influence of model selection (Figure~\ref{fig:label_efficiency}).
2D \epsSegpp is better than \epsSeg at every budget, by $0.159$ mean Dice at $80$ annotations and still by $0.052$ at $10{,}000$.
Read horizontally, the curves show a shift along the annotation axis rather than a constant offset: \epsSegpp reaches with $2{,}500$ annotations the Dice score ($0.768$) that \epsSeg needs about $9{,}000$ for.
Up to $600$ annotations, 2D \epsSegpp is never behind 3D \epsSeg while predicting a section about eleven times faster (Supp.\ Tab.~\ref{tab:runtime}); above roughly a thousand the volumetric input takes over.
Averaged over three seeds, 3D \epsSegpp is ahead of 3D \epsSeg at nine of twelve budgets, most clearly between $160$ and $1{,}200$ annotations ($+0.041$ to $+0.082$, Supp.\ Sec.~\ref{sec:label_efficiency_details}).
The differences between models shrink as the training budget approaches $160{,}000$ annotations.

\miniheadline{Dice Score as Selection Metric}
\epsSeg~\cite{epsseg} selected the checkpoint with the lowest total validation loss.
Selecting on the validation Dice score instead matters most where labels are scarce: pooled over all our models, it adds $0.143$ and $0.135$ mean Dice at $80$ and $160$ annotations, and makes no difference above about $20{,}000$ (Supp.\ Tab.~\ref{tab:selection-rules} and Supp.\ Fig.~\ref{fig:supp_label_efficiency}).
It also lifts 3D \epsSeg at $2{,}500$ annotations from $0.760$ to $0.829$ (Supp.\ Tab.~\ref{tab:abl-architecture}).
We therefore select on the Dice score throughout (Supp.\ Sec.~\ref{sec:training_details}).

\miniheadline{Validation Cost and Model Selection Strategies}
\label{sec:exp_selection}
\begin{figure*}[t!]
    \centering
    \includegraphics[width=\linewidth]{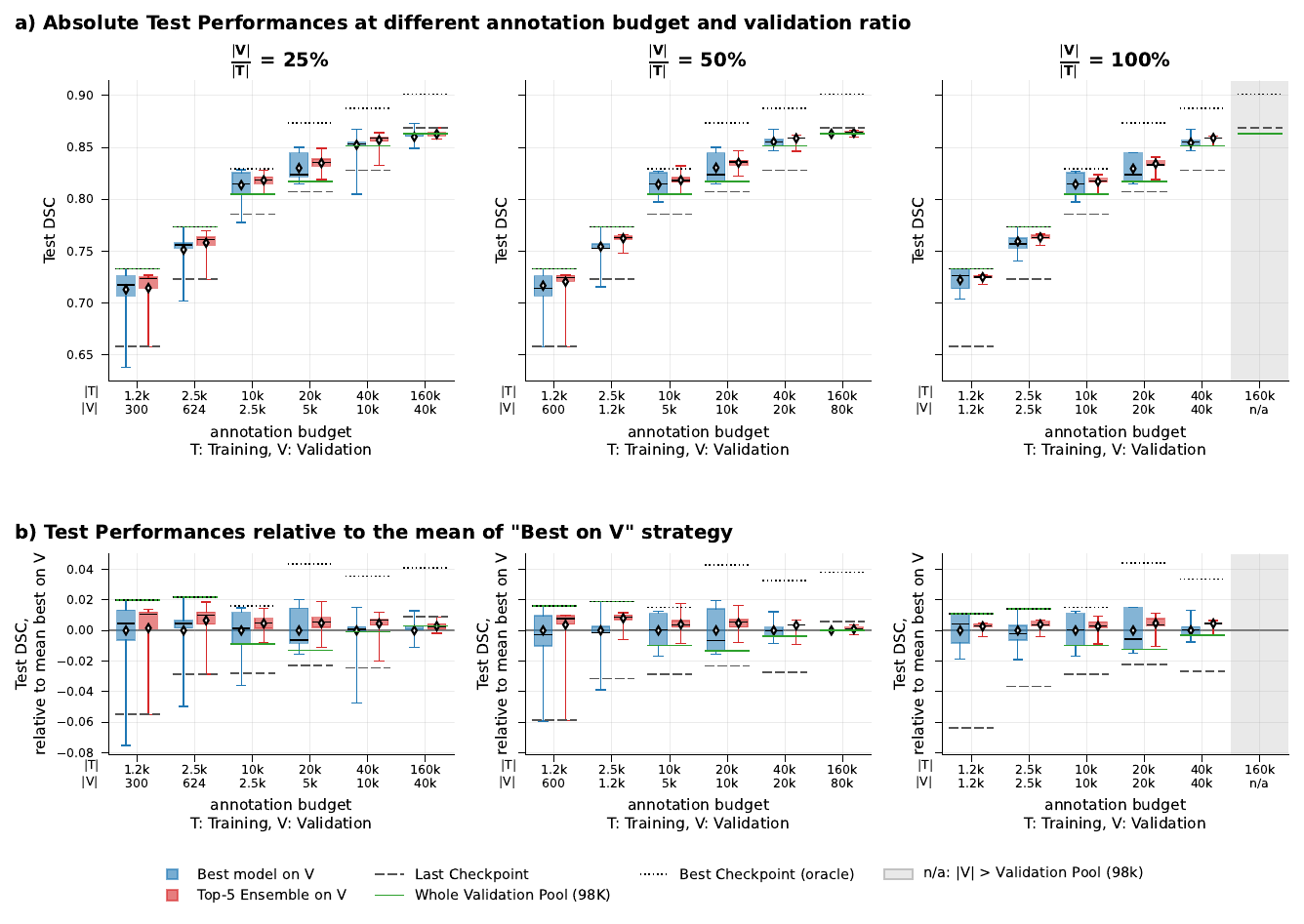}
    \caption{\textbf{Checkpoint selection at realistic validation budgets} (2D \epsSegpp, PoE).
    For each training budget, validation sets of $25\%$, $50\%$ and $100\%$ of $\lvert T\rvert$ are drawn $50$ times each.
    Blue: best checkpoint on $V$; red: top-$5$ ensemble on $V$; green: best checkpoint on the full pool of $98{,}000$ annotations; dashed: last checkpoint; dotted: test oracle.
    \textbf{(a)} Test Dice; \textbf{(b)} relative to the mean of the best-on-$V$ strategy. Supp.\ Fig.~\ref{fig:selection_strategies} shows more budgets and ratios.}
    \label{fig:selection_by_budget}
\end{figure*}

The label-efficiency experiment uses a validation pool far larger than most of its training budgets, which is unrealistic, and the selection rule matters most where labels are scarce.
We therefore ask how much of an annotation budget is worth spending on validation, and how the checkpoint should then be chosen.
For our 2D \epsSegpp, we compare the last checkpoint, which needs no validation labels, the single best checkpoint on $V$, and the top-$5$ ensemble of Section~\ref{sec:selection}, for validation sets of several sizes drawn $50$ times each (Figure~\ref{fig:selection_by_budget}; training details in Supp.\ Sec.~\ref{sec:training_details}).

For the smallest training budgets, the full validation pool selects the best model, as expected.
Between $5{,}000$ and $40{,}000$ training annotations, however, a smaller validation set is often the better choice: at $\lvert T\rvert = 20{,}000$, the best checkpoint on the full pool of $98{,}000$ annotations reaches $0.817$ mean Dice, the best on $5{,}000$ of them $0.830$, and the top-$5$ ensemble on the same $5{,}000$ $0.835$.
The ensemble and the single best checkpoint are close on average, but on the worst of the $50$ draws the ensemble is ahead in $72$ of $84$ combinations of training and validation budget, and in all of them whenever $\lvert V\rvert \le \lvert T\rvert/4$, by up to $0.19$ mean Dice (Supp.\ Sec.~\ref{sec:selection_details}).
The larger the training set, the less the selection strategy matters.

\section{Discussion}
\label{sec:discussion}

\miniheadline*{Multi-level experts or volumetric context}
The multi-level experts and the volumetric input are two routes to largely the same information, and at matched annotation most of the gain over 2D \epsSeg comes from the volumetric input (Table~\ref{tab:main-epssegpp-results}).
The experts gain substantially in 2D but add less on top of 3D inputs, mostly at low budgets (Figure~\ref{fig:label_efficiency}, Supp.\ Tab.~\ref{tab:abl-architecture}).
The choice between the two is therefore mainly one of cost.
Below about $600$ labels, the 2D experts deliver what 3D delivers, but at a fraction of the prediction time, and they are the only option for anisotropic or effectively two-dimensional data; with more labels, the volumetric input is worth its cost.
How the experts are trained also explains the choice of fusion rule.
Since none of them receives a loss of its own, they need not be individually reliable, so a product of experts, which adds their evidence in logit space, suits them better than averaging their distributions, which treats each expert as a calibrated classifier.
For the same reason, what a single level predicts on its own is not well defined; Supp.\ Sec.~\ref{sec:per_level} illustrates this.

\miniheadline*{Validation labels are annotation, too}
In the sparse regime, the rule that turns validation labels into a model matters as much as the architecture: at $80$ labels, selecting on the validation Dice score instead of the total loss is worth about as much as the multi-level experts (Supp.\ Tab.~\ref{tab:selection-rules}).
More validation labels are also not always better, and a small validation set combined with checkpoint ensembling is often the safer investment (Section~\ref{sec:exp_selection}).
A plausible explanation is that validation and test are different populations, so that beyond some size more validation labels only sharpen a model ranking that does not transfer to the test set.
A label-efficiency claim measured with a generous validation set therefore describes the architecture rather than the annotation a user will need, and we argue that sparsely supervised methods must report validation labels as part of their annotation cost.

\miniheadline*{Limitations and outlook}
To be comparable to prior work~\cite{epsseg}, we evaluate on a single dataset and test section, with three to five seeds and splits that are leakage-free by construction; testing whether the 2D--3D crossover and the validation-budget behaviour transfer to other vEM data, and how much a naive split would inflate results, are natural next steps.
The study of realistic validation budgets covers our 2D model only.
The main practical cost is prediction: classifying every voxel from its own patch is far slower than dense segmentation (Supp.\ Tab.~\ref{tab:runtime}), and predicting a sparse grid of voxels or distilling the ensemble into a dense network are natural remedies.
Our annotations are random, class-balanced voxels, an idealisation of the clicks a human annotator would place.
A cheap active-learning loop could close this gap: disagreement within the checkpoint ensemble or in the posterior of the hierarchical VAE marks voxels where a label is likely informative, and the predicted class probabilities can steer queries towards balanced classes.

\section{Conclusion}
\label{sec:conclusion}
Sparse annotation is only as efficient as the pipeline that consumes it, including the labels spent on validation and the rule that turns them into a model.
\epsSegpp ensembles twice, across the levels of a hierarchical VAE and across training time, and a leakage-free split makes its 2D and 3D variants comparable.
The outcome is a choice that follows the budget: multi-level experts in 2D when labels are few, volumetric inputs when they are not, and checkpoint ensembling on whatever validation set can be afforded.

\section*{LLM Usage Statement}
We used an LLM-based assistant (Claude, Anthropic) for coding and writing.
\emph{Code and experiments:} writing and debugging training, evaluation and plotting scripts, managing cluster jobs, and cross-checking reported numbers against experiment outputs.
\emph{Figures:} plotting code; the microscopy images and segmentations shown are unedited data and outputs of our models, and no figure content is AI-generated.
\emph{Main text and supplement:} drafting and editing of text, grammar, literature search, and verification of bibliography entries.
\emph{Ideation:} no research ideas; only suggestions for how to frame some aspects of the paper clearly within limited space.
The authors designed all experiments, verified all content, and take full responsibility for it.

{
    \small
    \bibliographystyle{ieeenat_fullname}
    \bibliography{main}
}

\clearpage

\maketitlesupplementary
\setcounter{section}{0}
\renewcommand{\thesection}{S\arabic{section}}
\renewcommand{\theHsection}{S\arabic{section}}
\setcounter{figure}{0}
\renewcommand{\thefigure}{S\arabic{figure}}
\renewcommand{\theHfigure}{S\arabic{figure}}
\setcounter{table}{0}
\renewcommand{\thetable}{S\arabic{table}}
\renewcommand{\theHtable}{S\arabic{table}}

\section{Baseline Details}
\label{sec:baselines_details}

\miniheadline*{Matching the annotation}
\epsSeg and \epsSegpp annotate individual voxels, whereas AGMM-SASS~\cite{agmm} and D3PM~\cite{d3pm} require at least one labelled voxel per class present in each input patch.
Starting from $\lvert T\rvert = 10{,}000$ training and $\lvert V\rvert = 6{,}124$ validation patch centres, the latter chosen to keep the validation-to-training ratio of our full annotation pool ($0.6125$), we therefore label one additional voxel per class present in the patch built around each patch centre.
This brings the same $10{,}000$ training patches to $25{,}822$ annotated voxels and the validation patches to $16{,}069$, and the baselines are trained on these patches.
In one patch the only mitochondria lie within the border margin, where no input patch can be centred, so that annotation cannot enter our training set; the baselines' training set therefore holds $25{,}823$ voxels and their total $41{,}892$, one more than ours in each case.
We then compare three configurations, each holding a different quantity fixed:
(i) the same annotation as the baselines, \ie, all $41{,}891$ expanded voxels;
(ii) the same training annotation as (i), but validated on the $6{,}124$ original patch centres alone, for a total of $31{,}946$ voxels;
(iii) the same patches as the baselines, but annotated only at their centres, \ie, the $16{,}124$ original patch centres.

\miniheadline*{Image data read by each method}
The methods do not read the same image data, even when they receive the same annotation.
In configuration (iii), our 2D models read exactly the baselines' $64\times64$ patches, while our 3D models read a $32^3$ neighbourhood around the same centres.
In configurations (i) and (ii), our models build one patch around every annotated voxel, and therefore read $25{,}822$ partially overlapping patches where the baselines read the original $10{,}000$.
What the comparison matches is the annotation, which is the quantity that domain experts have to be paid for, and not the amount of unlabelled image data each method reads.

\miniheadline*{Further observations}
At matched annotation, 3D \epsSegpp gains $0.092$ Dice over D3PM on granules, the smallest and most numerous structure, and $0.025$ on the nucleus.
Every configuration we train, including 2D \epsSeg, is ahead of AGMM-SASS.
Prediction cost separates the two baselines rather than placing us between them: AGMM-SASS segments densely and remains far cheaper at inference than any of our configurations, whereas D3PM is slower than all of them (Supp.\ Tab.~\ref{tab:runtime}).

\miniheadline*{Provenance of the baseline numbers}
We trained both baselines ourselves, using the authors' official implementations at commits \texttt{6a60ea0} (AGMM-SASS) and \texttt{849d390} (D3PM) and adding only the adapters that read BetaSeg and our annotation files; neither method is otherwise modified.
Both receive the annotation described above, \ie, the same $10{,}000$ training and $6{,}124$ validation patches and the same per-class expansion as our models.
For AGMM-SASS we keep the authors' Pascal VOC settings for the backbone, the dilation rates and the head learning-rate multiplier, \ie, a ResNet-101 initialised from ImageNet, dilations $[6,12,18]$ and a $\times 10$ multiplier on the head, and adapt the remaining optimisation settings to the $64 \times 64$ patches: stochastic gradient descent with learning rate $0.002$, momentum $0.9$, weight decay $10^{-4}$, batches of $128$ patches and $100$ epochs.
For D3PM we keep the authors' settings as they stand: $200$ diffusion steps, a focal loss with $\gamma = 2$, Adam with learning rate $10^{-4}$, batches of $32$ patches and $70$ epochs.
We re-score both methods on our common region with our own metric rather than quoting published numbers.
On each method's own region the re-scoring reproduces its reported per-class and mean Dice exactly, which is the control that the region is the only thing that changes; restricting to the common region then raises their mean Dice by $0.003$ and $0.002$ respectively.

\section{Volumetric Model Details}
\label{sec:volumetric_details}
For 3D inputs, the input patch becomes a cube instead of a square, and with it every spatial operator of the backbone, \ie, all convolutions, normalisations and resampling operators are replaced by their volumetric counterparts.
The rest of the architecture is unchanged: the number of hierarchy levels, their latent dimensions, the number of filters per level, the size of the masked region, and all loss terms and their weights are identical in the two configurations.
Since the spatial size of the features at each level is determined by the input size, the input of the semantic experts changes from $[32,16,8]$ in 2D to $[16,8,4]$ in 3D, and from $8\times8$ to $4^3$ for the single head of \epsSeg.
The volumetric input is not simply a larger context: at the $16$\,nm isotropic resolution of BetaSeg, a $64^2$ patch covers $1.02\times1.02\,\mu$m of a single section, whereas a $32^3$ patch covers $0.51\,\mu$m along each axis.

\section{Leakage-Free Data Preparation}
\label{sec:suppl_preprocessing}

\begin{table}[t]
\centering
\footnotesize
\setlength{\tabcolsep}{1.8pt}
\begin{tabular}{l!{\color{gray!50}\vrule}cccc!{\color{gray!50}\vrule}c}
\hline
Split & U & N & G & M & Avg. \\
\hline
\epsSeg, RLF~1~\cite{epsseg} & $0.79$ & $0.95$ & $0.69$ & $0.69$ & $0.78$ \\
\rowcolor{gray!15}
Ours, $10{,}000$ annotations & $0.798$ & $0.943$ & $0.633$ & $0.760$ & $0.783$ \\
$\Delta$ & $+0.008$ & $-0.007$ & $-0.057$ & $+0.070$ & $+0.003$ \\
\hline
\end{tabular}
\caption{\textbf{Our split reproduces the published \epsSeg score.}
2D \epsSeg as published~\cite{epsseg} and retrained under our sub-volume split with $10{,}000$ training annotations, scored on the test section.
Mean Dice agrees within $0.003$; per-class differences reach $0.07$.
RLF, the Relative Labeling Factor of \epsSeg~\cite{epsseg}, scales how many annotated voxels are drawn per class and section; RLF~$1$, its lowest reported setting, amounts to $10{,}029$ training voxels, so the two rows are at a matched budget.
Our row is a single 2D \epsSeg run, selected by the highest Dice score on its own validation set of $6{,}124$ annotations and scored on the common region, which is why it differs from the 2D \epsSeg row of Table~\ref{tab:main-epssegpp-results}, a top-$5$ ensemble trained on the expanded annotation.}
\label{tab:rlf_split}
\end{table}

Since \epsSegpp can take as input either a square or a cubic patch extracted from an input volume
and predict a segmentation for only the centre voxel, particular care must be taken to ensure the
correct separation between different data splits.
To prevent data leakage between training and validation splits when training 3D-input models and
promote a fair comparison between 2D and 3D-input models, we assigned spatial portions of the images to either training or validation areas.

Starting from a dataset of $N_v$ volumes of size $D \times H \times W$, we assigned
$N_{dev}$ volumes to the \emph{development set} and $N_{test}$ volumes to the test set.
We then split each volume in the development set into multiple sub-volumes of size
$S \times H \times W$, where $S \le D$ is the sub-volume depth.
Since the BetaSeg dataset contained a special label to mark areas that are outside of the ROI for
segmentation, we discarded every resulting sub-volume that is either smaller than $S$ along the
depth dimension, or contained only voxels outside of the ROI.
Let $\mathbf{P}_{\max} = [P_{\max,z}, P_{\max,y}, P_{\max,x}]$ denote the largest input patch size
used across our experiments, with $P_{\max,z} \le S$, and let $\mathbf{S} = [S,H,W]$ denote the
sub-volume size.
For each axis $i \in \{z,y,x\}$, let $m_i = \left\lfloor P_{\max,i}/2 \right\rfloor$ be the
corresponding half-patch margin.
We sampled patch centre coordinates only from positions that allow any patch with size up to
$\mathbf{P}_{\max}$ to be extracted without crossing image or sub-volume boundaries.
Formally, each centre coordinate $p_i$ in sub-volume space was required to satisfy
$m_i \le p_i < S_i - m_i $, where $S_i$ is the sub-volume size along axis $i$.
We then sampled $n_c$ coordinates per class $c\in\{1,\ldots,C\}$ within each valid sub-volume to
build the set of patch centres to be used for training and validation.
We consider the obtained sub-volumes as the atomic unit for splitting the development dataset in
training set $\lvert T\rvert$ and validation set $\lvert V\rvert$.
Therefore, we assign each sub-volume to either set by applying stratified K-Fold across the
original volume indices, to ensure that each development volume contributes equally to each split,
while we use the $K$ parameter to control the relative amount of training and validation data.
Finally, the assignments between coordinates and the respective sub-volumes, splits and folds are
stored to be re-used across experiments.
During training, we perform z-score normalisation using the statistics computed on the sub-volumes
assigned to the training set of the fold in use. During prediction, we apply the same statistics to
the input data.

By applying the pre-processing strategy above, we ensure that the 3D-input models never receive as
their input context a portion of the data assigned to another split. We tuned the parameter
$P_{max}$ to match the maximum input size of our experiments, and then balanced $S$ and $n_c$ based
on the dataset structure to ensure a good class representation while minimising data loss.

\section{Dataset and Evaluation Details}
\label{sec:dataset_details}
Of the seven semantic classes of BetaSeg~\cite{betaseg}, we keep nucleus, granules and mitochondria as in \epsSeg~\cite{epsseg} and merge the remaining four into the Unrecognised class; an additional non-semantic label marks the voxels outside the cell, which we exclude from every loss and metric.
Patch centres are sampled per class within each sub-volume.
Since the nucleus is not present in every sub-volume, we double its sampling density relative to the other classes, so that the resulting budgets are class-balanced.
Our models use no input padding, so they can only predict voxels whose full input patch lies inside the volume, and the 2D ($64^2$) and 3D ($32^3$) inputs reach different regions of the test section.
To make 2D and 3D results comparable, we score every experiment on the region both can predict, which contains $311{,}821$ voxels; this choice has a negligible effect on the performance of our 3D models (at most $0.002$ mean Dice, and in their favour).

The four classes partition this region, with the Unrecognised class occupying $177{,}187$ of its voxels ($56.8\%$), the granules $55{,}857$ ($17.9\%$), the nucleus $40{,}555$ ($13.0\%$) and the mitochondria $38{,}222$ ($12.3\%$).
They differ as much in how that mass is distributed as in its amount: the nucleus is a single connected component, the mitochondria are $25$ components of median area $836$ voxels, the granules $206$ components of median area $197$, and the Unrecognised class is one connected background of $177{,}171$ voxels beside $13$ isolated ones.
Per-class Dice has to be read with this in mind, \ie, the granules are the smallest and most numerous structure and the only class whose score is dominated by many small objects, whereas a single object decides the nucleus.
Components are counted within the region and with $8$-connectivity, so an object cut by its boundary counts once and one lying entirely in the discarded border ring does not count at all.

\section{Architecture Ablations}
\label{sec:ablation_details}
\begin{table*}[t]
\centering
\small
\setlength{\tabcolsep}{5pt}
\begin{tabular}{l!{\color{gray!50}\vrule}c!{\color{gray!50}\vrule}cc!{\color{gray!50}\vrule}cc}
\hline
 & \epsSeg & \epsSegpp PA & $\Delta$ PA, \epsSeg & \epsSegpp PoE & $\Delta$ PoE, \epsSeg \\
 & ($\pm$ SD) & ($\pm$ SD) & ($\pm$ SE) & ($\pm$ SD) & ($\pm$ SE) \\
\hline
\multicolumn{6}{l}{\textit{Validation pool, $\lvert V\rvert = 98{,}000$}} \\
2D best(loss)  & $0.794 \pm 0.011$ & $0.847 \pm 0.011$ & $+0.053 \pm 0.005$~{\scriptsize(5/5)} & $0.857 \pm 0.003$ & $+0.063 \pm 0.006$~{\scriptsize(5/5)} \\
2D best(CE)    & $0.801 \pm 0.009$ & $0.847 \pm 0.013$ & $+0.046 \pm 0.008$~{\scriptsize(5/5)} & $0.859 \pm 0.004$ & $+0.058 \pm 0.005$~{\scriptsize(5/5)} \\
2D best(dice)  & $0.804 \pm 0.008$ & $0.853 \pm 0.008$ & $+0.050 \pm 0.005$~{\scriptsize(5/5)} & $0.861 \pm 0.003$ & $+0.058 \pm 0.005$~{\scriptsize(5/5)} \\
3D best(loss)  & $0.877 \pm 0.008$ & $0.880 \pm 0.022$ & $+0.002 \pm 0.011$~{\scriptsize(2/3)} & $0.898 \pm 0.005$ & $+0.021 \pm 0.004$~{\scriptsize(3/3)} \\
3D best(CE)    & $0.888 \pm 0.006$ & $0.896 \pm 0.002$ & $+0.008 \pm 0.005$~{\scriptsize(2/3)} & $0.904 \pm 0.002$ & $+0.016 \pm 0.004$~{\scriptsize(3/3)} \\
3D best(dice)  & $0.905 \pm 0.004$ & $0.898 \pm 0.001$ & $-0.007 \pm 0.002$~{\scriptsize(0/3)} & $0.907 \pm 0.003$ & $+0.002 \pm 0.001$~{\scriptsize(3/3)} \\
\hline
\multicolumn{6}{l}{\textit{Common test region}} \\
2D best(loss)  & $0.651 \pm 0.017$ & $0.722 \pm 0.019$ & $+0.071 \pm 0.013$~{\scriptsize(5/5)} & $0.755 \pm 0.013$ & $+0.104 \pm 0.011$~{\scriptsize(5/5)} \\
2D best(CE)    & $0.660 \pm 0.012$ & $0.735 \pm 0.008$ & $+0.075 \pm 0.008$~{\scriptsize(5/5)} & $0.758 \pm 0.015$ & $+0.099 \pm 0.007$~{\scriptsize(5/5)} \\
2D best(dice)  & $0.653 \pm 0.013$ & $0.734 \pm 0.019$ & $+0.082 \pm 0.012$~{\scriptsize(5/5)} & $0.764 \pm 0.010$ & $+0.111 \pm 0.006$~{\scriptsize(5/5)} \\
3D best(loss)  & $0.760 \pm 0.017$ & $0.776 \pm 0.042$ & $+0.016 \pm 0.014$~{\scriptsize(2/3)} & $0.806 \pm 0.027$ & $+0.046 \pm 0.014$~{\scriptsize(3/3)} \\
3D best(CE)    & $0.779 \pm 0.021$ & $0.812 \pm 0.012$ & $+0.033 \pm 0.014$~{\scriptsize(3/3)} & $0.829 \pm 0.015$ & $+0.050 \pm 0.018$~{\scriptsize(3/3)} \\
3D best(dice)  & $0.829 \pm 0.009$ & $0.813 \pm 0.007$ & $-0.016 \pm 0.009$~{\scriptsize(0/3)} & $0.810 \pm 0.027$ & $-0.019 \pm 0.020$~{\scriptsize(1/3)} \\
\hline
\end{tabular}
\caption{\textbf{Effect of the architecture.}
All models are trained with a budget of $\lvert T\rvert = 2{,}500$ and $\lvert V\rvert = 98{,}000$.
Columns 4 and 6 report the difference between the runs, paired per seed; the number in parentheses is the number of seeds with a positive difference.
The upper block reports the Dice score on the validation pool and the lower one the Dice score on the common test region; the comparison is otherwise identical.}
\label{tab:abl-architecture}
\end{table*}

\begin{table*}[t]
\centering
\small
\setlength{\tabcolsep}{6pt}
\begin{tabular}{l!{\color{gray!50}\vrule}l!{\color{gray!50}\vrule}cc!{\color{gray!50}\vrule}c!{\color{gray!50}\vrule}c}
\hline
Feature size & Metric & PA & PoE & Difference & Seed agreement \\
\hline
\multicolumn{6}{l}{\textit{Validation pool, $\lvert V\rvert = 98{,}000$}} \\
2D 25\,\% & best(loss)  & $0.855 \pm 0.006$ & $0.869 \pm 0.004$ & $+0.014 \pm 0.002$ & 5/5 \\
 & best(CE)    & $0.860 \pm 0.003$ & $0.871 \pm 0.006$ & $+0.011 \pm 0.002$ & 5/5 \\
 & best(dice)  & $0.863 \pm 0.004$ & $0.872 \pm 0.005$ & $+0.010 \pm 0.002$ & 5/5 \\
2D 50\,\% & best(loss)  & $0.852 \pm 0.007$ & $0.868 \pm 0.003$ & $+0.016 \pm 0.003$ & 5/5 \\
 & best(CE)    & $0.855 \pm 0.005$ & $0.868 \pm 0.002$ & $+0.014 \pm 0.002$ & 5/5 \\
 & best(dice)  & $0.861 \pm 0.004$ & $0.869 \pm 0.002$ & $+0.008 \pm 0.001$ & 5/5 \\
2D 100\,\% & best(loss)  & $0.847 \pm 0.011$ & $0.857 \pm 0.003$ & $+0.010 \pm 0.004$ & 4/5 \\
 & best(CE)    & $0.847 \pm 0.013$ & $0.859 \pm 0.004$ & $+0.012 \pm 0.005$ & 4/5 \\
 & best(dice)  & $0.853 \pm 0.008$ & $0.861 \pm 0.003$ & $+0.008 \pm 0.003$ & 3/5 \\
3D 100\,\% & best(loss)  & $0.880 \pm 0.022$ & $0.898 \pm 0.005$ & $+0.019 \pm 0.010$ & 3/3 \\
 & best(CE)    & $0.896 \pm 0.002$ & $0.904 \pm 0.002$ & $+0.008 \pm 0.000$ & 3/3 \\
 & best(dice)  & $0.898 \pm 0.001$ & $0.907 \pm 0.003$ & $+0.009 \pm 0.002$ & 3/3 \\
\hline
\multicolumn{6}{l}{\textit{Common test region}} \\
2D 25\,\% & best(loss)  & $0.720 \pm 0.028$ & $0.754 \pm 0.021$ & $+0.034 \pm 0.008$ & 5/5 \\
 & best(CE)    & $0.728 \pm 0.028$ & $0.762 \pm 0.028$ & $+0.034 \pm 0.005$ & 5/5 \\
 & best(dice)  & $0.729 \pm 0.040$ & $0.765 \pm 0.027$ & $+0.036 \pm 0.010$ & 5/5 \\
2D 50\,\% & best(loss)  & $0.726 \pm 0.018$ & $0.753 \pm 0.025$ & $+0.028 \pm 0.006$ & 5/5 \\
 & best(CE)    & $0.744 \pm 0.027$ & $0.752 \pm 0.028$ & $+0.008 \pm 0.014$ & 4/5 \\
 & best(dice)  & $0.755 \pm 0.020$ & $0.757 \pm 0.022$ & $+0.002 \pm 0.007$ & 2/5 \\
2D 100\,\% & best(loss)  & $0.722 \pm 0.019$ & $0.755 \pm 0.013$ & $+0.033 \pm 0.009$ & 5/5 \\
 & best(CE)    & $0.735 \pm 0.008$ & $0.758 \pm 0.015$ & $+0.024 \pm 0.007$ & 5/5 \\
 & best(dice)  & $0.734 \pm 0.019$ & $0.764 \pm 0.010$ & $+0.029 \pm 0.008$ & 5/5 \\
3D 100\,\% & best(loss)  & $0.776 \pm 0.042$ & $0.806 \pm 0.027$ & $+0.030 \pm 0.021$ & 2/3 \\
 & best(CE)    & $0.812 \pm 0.012$ & $0.829 \pm 0.015$ & $+0.016 \pm 0.004$ & 3/3 \\
 & best(dice)  & $0.813 \pm 0.007$ & $0.810 \pm 0.027$ & $-0.003 \pm 0.012$ & 1/3 \\
\hline
\end{tabular}
\caption{\textbf{Effect of the fusion rule: posterior averaging (PA) vs.\ Product of Experts (PoE) at equal feature size.}
$\lvert T\rvert = 2{,}500$ training and $\lvert V\rvert = 98{,}000$ validation annotations; $5$ seeds in 2D, $3$ in 3D; mean Dice on the common region of the test section.
Within a row block only the fusion rule changes.
Score columns are mean $\pm$ SD over seeds; the difference is paired per seed (mean $\pm$ SEM), and seed agreement counts the seeds in which PoE is ahead.
PoE is at least as good as PA in $11$ of $12$ comparisons, and the single exception ($-0.003 \pm 0.012$) is within noise.
The upper block reports the Dice score on the validation pool and the lower one the Dice score on the common test region; the comparison is otherwise identical.}
\label{tab:abl-aggregation}
\end{table*}

\begin{table*}[t]
\centering
\small
\setlength{\tabcolsep}{5pt}
\begin{tabular}{l!{\color{gray!50}\vrule}l!{\color{gray!50}\vrule}ccc!{\color{gray!50}\vrule}cc}
\hline
 & Metric & 25\,\% $[8,4,2]$ & 50\,\% $[16,8,4]$ & 100\,\% $[32,16,8]$ & $25 \rightarrow 50$ & $50 \rightarrow 100$ \\
\hline
\multicolumn{7}{l}{\textit{Validation pool, $\lvert V\rvert = 98{,}000$}} \\
PA & best(loss)  & $0.855 \pm 0.006$ & $0.852 \pm 0.007$ & $0.847 \pm 0.011$ & $-0.003 \pm 0.005$~{\scriptsize(1/5)} & $-0.004 \pm 0.007$~{\scriptsize(2/5)} \\
 & best(CE)    & $0.860 \pm 0.003$ & $0.855 \pm 0.005$ & $0.847 \pm 0.013$ & $-0.005 \pm 0.001$~{\scriptsize(0/5)} & $-0.008 \pm 0.006$~{\scriptsize(1/5)} \\
 & best(dice)  & $0.863 \pm 0.004$ & $0.861 \pm 0.004$ & $0.853 \pm 0.008$ & $-0.002 \pm 0.002$~{\scriptsize(2/5)} & $-0.007 \pm 0.004$~{\scriptsize(2/5)} \\
PoE & best(loss)  & $0.869 \pm 0.004$ & $0.868 \pm 0.003$ & $0.857 \pm 0.003$ & $-0.002 \pm 0.002$~{\scriptsize(1/5)} & $-0.010 \pm 0.002$~{\scriptsize(0/5)} \\
 & best(CE)    & $0.871 \pm 0.006$ & $0.868 \pm 0.002$ & $0.859 \pm 0.004$ & $-0.002 \pm 0.003$~{\scriptsize(2/5)} & $-0.009 \pm 0.002$~{\scriptsize(0/5)} \\
 & best(dice)  & $0.872 \pm 0.005$ & $0.869 \pm 0.002$ & $0.861 \pm 0.003$ & $-0.004 \pm 0.002$~{\scriptsize(1/5)} & $-0.007 \pm 0.002$~{\scriptsize(0/5)} \\
\hline
\multicolumn{7}{l}{\textit{Common test region}} \\
PA & best(loss)  & $0.720 \pm 0.028$ & $0.726 \pm 0.018$ & $0.722 \pm 0.019$ & $+0.006 \pm 0.019$~{\scriptsize(2/5)} & $-0.003 \pm 0.010$~{\scriptsize(3/5)} \\
 & best(CE)    & $0.728 \pm 0.028$ & $0.744 \pm 0.027$ & $0.735 \pm 0.008$ & $+0.016 \pm 0.011$~{\scriptsize(4/5)} & $-0.009 \pm 0.014$~{\scriptsize(2/5)} \\
 & best(dice)  & $0.729 \pm 0.040$ & $0.755 \pm 0.020$ & $0.734 \pm 0.019$ & $+0.027 \pm 0.017$~{\scriptsize(4/5)} & $-0.021 \pm 0.008$~{\scriptsize(0/5)} \\
PoE & best(loss)  & $0.754 \pm 0.021$ & $0.753 \pm 0.025$ & $0.755 \pm 0.013$ & $-0.001 \pm 0.017$~{\scriptsize(3/5)} & $+0.002 \pm 0.013$~{\scriptsize(3/5)} \\
 & best(CE)    & $0.762 \pm 0.028$ & $0.752 \pm 0.028$ & $0.758 \pm 0.015$ & $-0.011 \pm 0.019$~{\scriptsize(2/5)} & $+0.007 \pm 0.017$~{\scriptsize(2/5)} \\
 & best(dice)  & $0.765 \pm 0.027$ & $0.757 \pm 0.022$ & $0.764 \pm 0.010$ & $-0.008 \pm 0.018$~{\scriptsize(2/5)} & $+0.006 \pm 0.012$~{\scriptsize(2/5)} \\
\hline
\end{tabular}
\caption{\textbf{Effect of the feature crop that reaches the experts} (2D only).
$\lvert T\rvert = 2{,}500$ training and $\lvert V\rvert = 98{,}000$ validation annotations, $5$ seeds, mean Dice on the common region of the test section.
Percentages are fractions of the native per-level feature sizes; the bracketed triples are the resulting spatial sizes per level.
Score columns are mean $\pm$ SD; the two transition columns are paired per seed (mean $\pm$ SEM), with the number of seeds that improve in parentheses.
Under PoE, every transition is within about one SEM of zero; under PA, the $50\%$ crop is best when selecting on Dice.
The upper block reports the Dice score on the validation pool and the lower one the Dice score on the common test region; the comparison is otherwise identical.}
\label{tab:abl-feature-size}
\end{table*}

We compare the architectural choices we contribute, namely the multi-level experts, the fusion rule, and the cropping of the features that reach the experts.
Preliminary experiments showed that this comparison is sensitive to the training budget $\lvert T\rvert$, to selection noise from a small validation set, and to the selection metric.
Since our aim is an architecture that performs well with little annotation, we use a training budget of $\lvert T\rvert = 2{,}500$ annotations and the full validation pool of $\lvert V\rvert = 98{,}000$, train $5$ seeds in 2D and $3$ in 3D, and select checkpoints with early stopping under three metrics: total validation loss, cross-entropy, and Dice.
Every comparison below is made on both signals, the Dice score on the validation pool and on the common test region; each table carries the validation block first and the test block beneath it.
The two agree on every choice we adopt; the one setting they do not separate is the 3D feature crop, where neither supports a change from the unmodified architecture.
Validation Dice runs $0.119$ (2D) and $0.092$ (3D) above test Dice and compresses the differences between model variants to roughly half the margin, while ranking them in the same order ($\rho = 0.71$ to $0.79$ in 2D and $1.00$ in 3D under the loss and cross-entropy metrics); the exception is 3D under the Dice metric, where the variants lie within $0.019$ mean Dice on test and neither ranking is meaningful.

\miniheadline*{Multi-level experts}
For the 2D models, placing an expert at every level improves over \epsSeg on validation in all six comparisons of Supp.\ Tab.~\ref{tab:abl-architecture}, \ie, for both fusion rules and under each selection metric, and in every case for all five seeds, by $+0.042$ to $+0.063$ mean Dice.
Test agrees and at a wider margin: posterior averaging gains between $+0.071$ and $+0.082$ and the Product of Experts between $+0.099$ and $+0.111$, bringing the 2D model from $0.653$ to $0.764$ under the Dice metric.
The advantage therefore depends neither on how the checkpoint is selected nor on which signal it is read from.
For the 3D models the change is equivocal on both: every validation difference lies within $\pm 0.021$, and on test the gain depends on the metric and on the fusion rule: under the validation loss and the cross-entropy, posterior averaging gains $+0.016$ and $+0.033$ and the Product of Experts $+0.046$ and $+0.050$, while under the Dice score \epsSeg is ahead of them by $0.016$ and $0.019$ respectively.
The 3D baseline is already strong at this budget ($0.829$), \ie, the volumetric input appears to supply part of the information that the experts recover in 2D.

\miniheadline*{Fusion rule}
Supp.\ Tab.~\ref{tab:abl-aggregation} compares the fusion rules at each feature-crop size.
On validation the Product of Experts is at least as good as posterior averaging in all $12$ comparisons.
On test it leads in $11$ of the $12$ experiments, by $+0.002$ to $+0.036$ mean Dice.
The single exception is the 3D model with the full feature map under the Dice metric, where posterior averaging is ahead by $0.003 \pm 0.012$ on test, within noise, while the Product of Experts is ahead by $0.009$ on validation for all three seeds.

\begin{table*}[t]
\centering
\small
\setlength{\tabcolsep}{5pt}
\begin{tabular}{l!{\color{gray!50}\vrule}cc!{\color{gray!50}\vrule}cc!{\color{gray!50}\vrule}cc}
\hline
 & \multicolumn{2}{c}{experts: 3D \epsSegpp $-$ 3D \epsSeg} & \multicolumn{2}{c}{crop: 2D $50\,\%$ $-$ $100\,\%$} & \multicolumn{2}{c}{crop: 3D $50\,\%$ $-$ $100\,\%$} \\
$\lvert T\rvert$ & val & test & val & test & val & test \\
\hline
80        & $-0.043$ & $-0.017$ & $+0.016$ & $+0.010$ & $+0.062$ & $+0.047$ \\
160       & $+0.008$ & $+0.041$ & $+0.038$ & $+0.058$ & $+0.020$ & $+0.014$ \\
300       & $+0.016$ & $+0.053$ & $+0.018$ & $+0.011$ & $-0.001$ & $-0.011$ \\
600       & $+0.029$ & $+0.082$ & $+0.027$ & $-0.002$ & $-0.008$ & $-0.034$ \\
$1{,}200$ & $+0.012$ & $+0.043$ & $+0.012$ & $+0.012$ & $+0.001$ & $-0.011$ \\
$2{,}500$ & $+0.002$ & $-0.019$ & $+0.009$ & $+0.007$ & $+0.001$ & $+0.030$ \\
5k        & $+0.008$ & $+0.013$ & $+0.002$ & $-0.005$ & $+0.002$ & $-0.029$ \\
10k       & $+0.004$ & $+0.013$ & $+0.002$ & $+0.012$ & $+0.001$ & $+0.005$ \\
20k       & $+0.005$ & $-0.006$ & $+0.001$ & $+0.003$ & $-0.001$ & $-0.002$ \\
40k       & $+0.000$ & $+0.005$ & $+0.001$ & $+0.014$ & $+0.003$ & $+0.019$ \\
80k       & $+0.003$ & $+0.015$ & $+0.001$ & $+0.019$ & $+0.000$ & $-0.012$ \\
160k      & $+0.006$ & $+0.012$ & $+0.001$ & $+0.001$ & $-0.006$ & $-0.002$ \\
\hline
\end{tabular}
\caption{\textbf{Effect of the multi-level experts and of the feature crop at every training budget.} Each column pair is one paired comparison, reported as the Dice score of the first variant minus that of the second and averaged over the three seeds, so a positive value favours the first: 3D \epsSegpp against 3D \epsSeg, then the $50\,\%$ feature crop against the full feature map, both within \epsSegpp, in 2D and in 3D. The two variants of a pair share the training budget $\lvert T\rvert$ and the seed and differ only in the setting compared. Scores are read on the validation pool ($\lvert V\rvert = 98{,}000$ annotations) and on the common test region, on the runs of the label-efficiency experiment (Figure~\ref{fig:label_efficiency}); each run contributes the single checkpoint with the highest validation Dice score, so the validation columns are optimistic for both variants alike.}
\label{tab:budget-sweep}
\end{table*}

\miniheadline*{Feature cropping}
Supp.\ Tab.~\ref{tab:abl-feature-size} compares the crop sizes within each fusion rule at a single budget, where neither signal separates the sizes: the differences are small relative to the seed spread, most of them within one standard error of zero, and validation favours the smallest crop at every step while test points different ways depending on the fusion rule.
The label-efficiency experiment (Figure~\ref{fig:label_efficiency}) repeats the comparison at twelve budgets, reported in Supp.\ Tab.~\ref{tab:budget-sweep}, and does separate them.
In 2D the $50\%$ crop is ahead of the full feature map at all twelve budgets on validation, and at least as good at ten of the twelve on test, the two exceptions costing $0.002$ and $0.005$ mean Dice.
In 3D neither signal gives a consistent winner: on validation the two sizes lie within $0.005$ mean Dice at eight of the twelve budgets and the remaining four split evenly between them; on test the differences are larger and change sign repeatedly, from $+0.047$ at $80$ annotations to $-0.034$ at $600$, with the crop ahead at five budgets.
We therefore keep the experts' feature map at $[16,8,4]$ in both configurations: a $50\%$ crop of the natural map in 2D, and the natural map itself in 3D, whose $32^3$ input already yields those sizes (Supp.\ Sec.~\ref{sec:volumetric_details}).
The experiments constrain this choice from above but not from below: doubling the extent in 2D, to $[32,16,8]$, is worse at ten of the twelve budgets, whereas halving it in 3D, to $[8,4,2]$, changes nothing consistently.

\section{Label-Efficiency Details}
\label{sec:label_efficiency_details}
Each training set of the label-efficiency experiment (Figure~\ref{fig:label_efficiency}) is drawn from the full training pool by sampling an equal number of annotations per class, roughly halving the number of annotations at each step, so that all training sets are nested.
At the lowest budgets, 2D \epsSegpp matches the 3D \epsSeg baseline: up to $600$ training annotations it is never behind it, by $+0.077$ mean Dice at $80$ and $+0.016$ at $600$.
The shift along the annotation axis described in Section~\ref{sec:exp_label_efficiency} holds at higher budgets too: with $10{,}000$ annotations, 2D \epsSegpp reaches the Dice score that 2D \epsSeg needs about $40{,}000$ for.
Above roughly a thousand annotations the volumetric input takes over, and averaged over three seeds 3D \epsSegpp is ahead of 3D \epsSeg at nine of the twelve budgets, most clearly between $160$ and $1{,}200$ annotations, where it gains between $+0.041$ and $+0.082$ mean Dice.
Supp.\ Tab.~\ref{tab:budget-sweep} gives the difference at every budget on both signals; on validation only the smallest budget, $80$ annotations, favours \epsSeg.
The multi-level experts thus recover in 2D part of the information that the volumetric input provides directly, and keep adding to it once that input is available.
Supp.\ Fig.~\ref{fig:supp_label_efficiency} repeats the experiment under the other selection rules.

\section{Training Details}
\label{sec:training_details}
We train with a batch size of $512$ in 2D and $96$ in 3D, drawing each batch class-balanced over the training annotations.
One pseudo-epoch is exactly $100$ optimiser steps at every training budget, so a budget changes which annotations are seen but not how long an epoch is; epochs are therefore a checkpoint-sampling interval rather than a pass over the data.
Every patch in a batch is centred on an annotated voxel: all losses, the inpainting term included, are computed on these training patches, and no patch centred on an unlabelled location is ever used.
All models were trained on a single NVIDIA Tesla V100-SXM2 with $32$\,GB, which is also the GPU of every timing in Supp.\ Tab.~\ref{tab:runtime}.

We use the Adamax optimiser with a learning rate of $3 \times 10^{-5}$, no weight decay, automatic mixed precision, and gradient-norm clipping at $1.0$.

The learning-rate schedule is the only optimisation setting that differs between our experiments.
The architecture selection and the label-efficiency experiments use the schedule of \epsSeg, \ie, ReduceLROnPlateau on the validation loss with patience $10$, factor $0.9$ and floor $10^{-12}$, stepped once per validation epoch of $100$ training batches.
The validation-budget study and the final models compared with the baselines use a fixed schedule instead, so that training does not depend on the validation set.
We obtained it by fitting an exponential decay to the median learning-rate trajectory of the label-efficiency runs (Figure~\ref{fig:label_efficiency}), whose ReduceLROnPlateau schedule was driven by the validation loss on the full validation pool:
\begin{equation}
  \eta(t) = \eta_0\,\gamma^{\max(0,\,t - H)/P},
  \label{eq:lr_schedule}
\end{equation}
with $\eta_0 = 3 \times 10^{-5}$, $\gamma = 0.9$, a hold of $H = 1{,}500$ and a decay period of $P = 2{,}750$, and $t$ counted in optimiser steps.
The learning rate is thus constant for the first $1{,}500$ steps and ends a $15K$-step run at $0.60\,\eta_0$.
The form has no floor, so no $\eta_{\min}$ is needed; one schedule is used for every budget and for both 2D and 3D.
At each optimiser step we took the median learning rate over the runs still training at that step, and fitted the decay to that curve.
Fitting instead to each run's final learning rate produces a decay that is much too fast, because a run that stops early has made the same reductions in fewer steps, so such a fit is dominated by the shortest runs, whereas every model trained with this schedule runs to a fixed length.
The shape of the schedule thus carries indirect information from the full validation pool, from which our smaller validation sets are also drawn.
No run that uses this schedule sees a validation signal of its own before checkpoint selection.

Supp.\ Tab.~\ref{tab:runtime} lists training, validation and prediction times.
Our models are also considerably smaller than the baselines: $3.8$M and $3.9$M trainable parameters for 2D \epsSeg and \epsSegpp, and $10.6$M and $11.5$M in 3D, against $59.5$M for AGMM-SASS and $46.6$M for D3PM.
\begin{table*}[t]
\centering
\small
\setlength{\tabcolsep}{6pt}
\begin{tabular}{l!{\color{gray!50}\vrule}r!{\color{gray!50}\vrule}rr!{\color{gray!50}\vrule}rr!{\color{gray!50}\vrule}rr!{\color{gray!50}\vrule}rr}
\hline
 & Trainable & \multicolumn{2}{c}{Training} & \multicolumn{2}{c}{Validation (s / ckpt)} & \multicolumn{2}{c}{Prediction, one slice} & \multicolumn{2}{c}{Total (h)} \\
Model & params & steps & h & $6{,}124$ & $16{,}069$ & time & rel. & $6{,}124$ & $16{,}069$ \\
\hline
\multicolumn{10}{l}{\textit{baselines: validation interleaved with training, one prediction}} \\
AGMM-SASS~\cite{agmm} & $59.5$M & $7{,}800$  & $3.6$ & \multicolumn{2}{c}{incl.} & $1.03$\,s & $<0.01\times$ & - & {$3.6$} \\
D3PM~\cite{d3pm}      & $46.6$M & $21{,}910$ & $2.3$ & \multicolumn{2}{c}{incl.} & $14.7$\,h & $304\times$ & - &{$17.0$} \\
\hline
\multicolumn{10}{l}{\textit{ours: all $150$ checkpoints validated post hoc, then prediction of the top-$5$ ensemble}} \\
2D \epsSeg            & $3.8$M  & $15{,}000$ & $5.6$  & $5$  & $10$  & $2.9$\,min  & $1\times$  & $6.0$  & $6.3$  \\
3D \epsSeg            & $10.6$M & $15{,}000$ & $10.7$ & $40$ & $102$ & $32.8$\,min & $11\times$ & $15.1$ & $17.7$ \\
\rowcolor{gray!15}
2D \epsSegpp          & $3.9$M  & $15{,}000$ & $5.9$  & $5$  & $11$  & $2.9$\,min  & $1\times$  & $6.4$  & $6.6$  \\
\rowcolor{gray!15}
3D \epsSegpp          & $11.5$M & $15{,}000$ & $10.8$ & $41$ & $105$ & $32.8$\,min & $11\times$ & $15.2$ & $17.9$ \\
\hline
\end{tabular}
\caption{\textbf{Model size and runtime.}
Baselines interleave validation with training, so their validation time is part of the training time.
Our models are trained for $15K$ steps, storing a checkpoint every $100$ steps; all $150$ checkpoints are validated \textit{post hoc} (seconds per checkpoint, for both validation-set sizes of Table~\ref{tab:main-epssegpp-results}).
Prediction time is for one section and a single model, and rel.\ is relative to 2D \epsSeg; the top-$5$ ensemble takes five times as long.
The total sums training, post-hoc validation and ensemble prediction of one section.
All timings are on a single NVIDIA Tesla V100-SXM2 with $32$\,GB.}
\label{tab:runtime}
\end{table*}

\begin{figure*}[t]
    \centering

    \vspace{0.6em}
    \begin{subfigure}[t]{\textwidth}
        \centering
        \includegraphics[width=\textwidth]{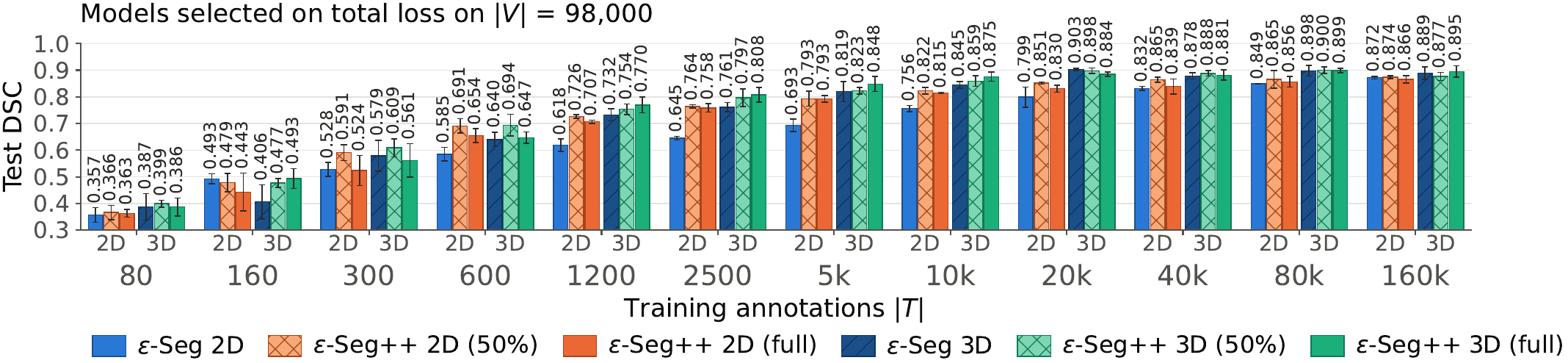}
        \caption{Models Selected on validation Total Loss}
        \label{fig:supp_label_efficiency_total_loss}
    \end{subfigure}

    \vspace{0.6em}
    \begin{subfigure}[t]{\textwidth}
        \centering
        \includegraphics[width=\textwidth]{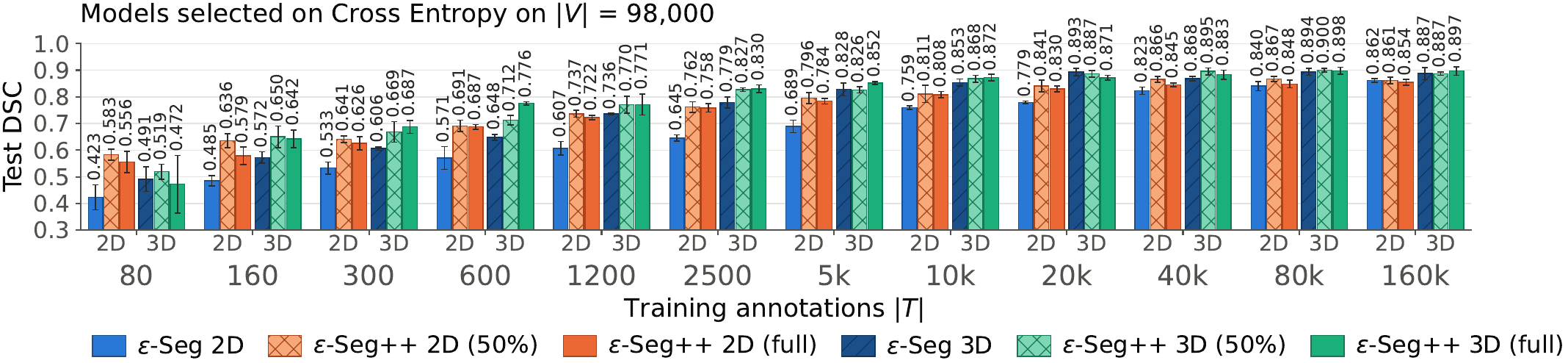}
        \caption{Models Selected on validation Cross Entropy term}
        \label{fig:supp_label_efficiency_ce}
    \end{subfigure}

    \begin{subfigure}[t]{\textwidth}
        \centering
        \includegraphics[width=\textwidth]{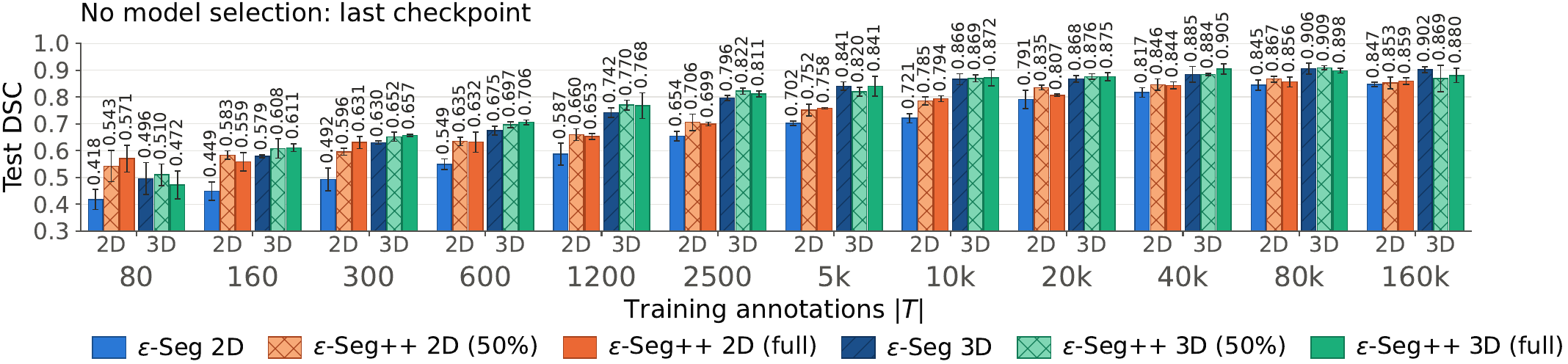}
        \caption{No selection, final checkpoint of every run}
        \label{fig:supp_label_efficiency_last}
    \end{subfigure}

    \caption{
        \textbf{Test performance of the architectures at each training budget, under three model-selection rules.} Each model is trained on a nested subset of the whole training pool, spanning $80 \rightarrow 160K$ annotations, and validated on the same validation pool, consisting of $98K$ annotated voxels. Error bars are computed using three different seeds.
        Training is performed using Early Stopping with $\text{patience}=20$ and monitoring the best validation Dice score, as we found that it led to longer training runs.
        For these experiments we decoupled the early-stopping criterion, \ie, when to stop training, from the metric used to select the best checkpoint.
        Figure~\textbf{(a)} reports the models according to their total validation loss, as used in \epsSeg \cite{epsseg}. Figure~\textbf{(b)} shows the best models according to only the validation Cross-Entropy term, while Figure~\textbf{(c)} shows the case in which no selection metric is used, i.e., we select the final checkpoint for each run.
        By comparing the different selection rules it is possible to notice that while at higher training budgets the models are ranked and perform similarly, at lower budgets the total loss penalises the model, as it is not able to select models that generalise to a different volume. For this reason we select on the Dice score throughout (Figure~\ref{fig:label_efficiency}), which we found to be comparable to the cross-entropy and never worse than it on average at any budget.
    }
    \label{fig:supp_label_efficiency}
\end{figure*}

\miniheadline*{Selection metric}
\epsSeg selected the checkpoint with the lowest total validation loss.
In the architecture ablation (Supp.\ Tab.~\ref{tab:abl-architecture}), the choice of metric has little effect on 2D \epsSeg ($0.009$ mean Dice), whereas selecting 3D \epsSeg on the validation Dice score raises its test performance from $0.760$ to $0.829$.
Supp.\ Fig.~\ref{fig:supp_label_efficiency} repeats the label-efficiency experiment with checkpoints selected by the total validation loss, by the cross-entropy, and without selection (last checkpoint), and Supp.\ Tab.~\ref{tab:selection-rules} gives the differences per model and budget.
Above roughly $20{,}000$ annotations the metric is essentially irrelevant, whereas at the smallest budgets it dominates: pooled over all models, the Dice score is worth $+0.143$ and $+0.135$ mean Dice over the validation loss at $80$ and $160$ annotations.
Against the cross-entropy the margin is much smaller but consistent, \ie, positive at all twelve budgets.
\begin{table*}[t]
\centering
\scriptsize
\setlength{\tabcolsep}{2.2pt}
\begin{tabular}{ll!{\color{gray!50}\vrule}rrrrrrrrrrrr!{\color{gray!50}\vrule}r}
\hline
Model & vs. & 80 & 160 & 300 & 600 & 1{,}200 & 2{,}500 & 5k & 10k & 20k & 40k & 80k & 160k & all \\
\hline
\multirow{3}{*}{2D \epsSeg} & last & $+0.006$ & $+0.052$ & $+0.047$ & $+0.043$ & $+0.037$ & $-0.003$ & $+0.001$ & $+0.055$ & $-0.005$ & $+0.012$ & $+0.010$ & $+0.020$ & $+0.023$ \\
 & best(loss) & $+0.067$ & $+0.009$ & $+0.011$ & $+0.007$ & $+0.005$ & $+0.006$ & $+0.010$ & $+0.020$ & $-0.013$ & $-0.002$ & $+0.006$ & $-0.005$ & $+0.010$ \\
 & best(CE) & $+0.001$ & $+0.016$ & $+0.006$ & $+0.021$ & $+0.017$ & $+0.006$ & $+0.014$ & $+0.016$ & $+0.007$ & $+0.006$ & $+0.014$ & $+0.005$ & $+0.011$ \\
\hline
\multirow{3}{*}{3D \epsSeg} & last & $+0.012$ & $+0.017$ & $+0.007$ & $0.000$ & $+0.014$ & $+0.034$ & $+0.002$ & $-0.005$ & $+0.019$ & $-0.012$ & $-0.011$ & $-0.018$ & $+0.005$ \\
 & best(loss) & $+0.121$ & $+0.190$ & $+0.058$ & $+0.036$ & $+0.024$ & $+0.069$ & $+0.023$ & $+0.017$ & $-0.016$ & $-0.006$ & $-0.003$ & $-0.004$ & $+0.042$ \\
 & best(CE) & $+0.016$ & $+0.023$ & $+0.031$ & $+0.027$ & $+0.021$ & $+0.050$ & $+0.015$ & $+0.008$ & $-0.006$ & $+0.004$ & $+0.001$ & $-0.003$ & $+0.016$ \\
\hline
\multirow{3}{*}{2D \epsSegpp 100\,\%} & last & $+0.003$ & $+0.019$ & $-0.003$ & $+0.060$ & $+0.072$ & $+0.062$ & $+0.044$ & $+0.022$ & $+0.041$ & $+0.006$ & $-0.008$ & $+0.002$ & $+0.027$ \\
 & best(loss) & $+0.211$ & $+0.135$ & $+0.104$ & $+0.038$ & $+0.018$ & $+0.003$ & $+0.010$ & $+0.001$ & $+0.017$ & $+0.011$ & $-0.008$ & $-0.005$ & $+0.044$ \\
 & best(CE) & $+0.018$ & $-0.001$ & $+0.002$ & $+0.005$ & $+0.002$ & $+0.003$ & $+0.018$ & $+0.008$ & $+0.017$ & $+0.005$ & $0.000$ & $+0.007$ & $+0.007$ \\
\hline
\multirow{3}{*}{2D \epsSegpp 50\,\%} & last & $+0.041$ & $+0.052$ & $+0.043$ & $+0.056$ & $+0.077$ & $+0.062$ & $+0.046$ & $+0.043$ & $+0.015$ & $+0.018$ & $0.000$ & $+0.009$ & $+0.038$ \\
 & best(loss) & $+0.217$ & $+0.157$ & $+0.048$ & $-0.001$ & $+0.010$ & $+0.004$ & $+0.004$ & $+0.006$ & $-0.001$ & $-0.001$ & $+0.002$ & $-0.012$ & $+0.036$ \\
 & best(CE) & $0.000$ & $0.000$ & $-0.002$ & $0.000$ & $0.000$ & $+0.006$ & $+0.002$ & $+0.017$ & $+0.009$ & $-0.002$ & $0.000$ & $+0.001$ & $+0.003$ \\
\hline
\multirow{3}{*}{3D \epsSegpp 100\,\%} & last & $+0.018$ & $+0.026$ & $+0.032$ & $+0.052$ & $+0.031$ & $0.000$ & $+0.014$ & $+0.003$ & $+0.005$ & $-0.027$ & $+0.011$ & $+0.016$ & $+0.015$ \\
 & best(loss) & $+0.105$ & $+0.144$ & $+0.129$ & $+0.111$ & $+0.029$ & $+0.004$ & $+0.007$ & $0.000$ & $-0.004$ & $-0.004$ & $+0.010$ & $+0.001$ & $+0.044$ \\
 & best(CE) & $+0.019$ & $-0.005$ & $+0.002$ & $-0.017$ & $+0.028$ & $-0.019$ & $+0.003$ & $+0.003$ & $+0.009$ & $-0.006$ & $+0.011$ & $-0.001$ & $+0.002$ \\
\hline
\multirow{3}{*}{3D \epsSegpp 50\,\%} & last & $+0.027$ & $+0.043$ & $+0.027$ & $+0.027$ & $+0.018$ & $+0.019$ & $+0.006$ & $+0.011$ & $+0.001$ & $+0.012$ & $-0.012$ & $+0.026$ & $+0.017$ \\
 & best(loss) & $+0.138$ & $+0.175$ & $+0.071$ & $+0.031$ & $+0.034$ & $+0.044$ & $+0.003$ & $+0.022$ & $-0.020$ & $+0.009$ & $-0.003$ & $+0.018$ & $+0.043$ \\
 & best(CE) & $+0.018$ & $+0.001$ & $+0.011$ & $+0.012$ & $+0.018$ & $+0.014$ & $0.000$ & $+0.013$ & $-0.009$ & $+0.001$ & $-0.002$ & $+0.008$ & $+0.007$ \\
\hline
\multicolumn{15}{l}{\textit{pooled over all six models; $\pm$ SE over the $216$ cells, with win/loss/tie counts}} \\
\hline
\multicolumn{2}{l}{best(dice) $-$ last} & $+0.018$ & $+0.035$ & $+0.025$ & $+0.040$ & $+0.041$ & $+0.029$ & $+0.019$ & $+0.022$ & $+0.013$ & $+0.002$ & $-0.002$ & $+0.009$ & $+0.021 \pm 0.002$ \\
\multicolumn{2}{l}{\quad\scriptsize W/L/T} & {\scriptsize 11/7/0} & {\scriptsize 18/0/0} & {\scriptsize 16/2/0} & {\scriptsize 16/2/0} & {\scriptsize 17/1/0} & {\scriptsize 14/4/0} & {\scriptsize 13/5/0} & {\scriptsize 15/3/0} & {\scriptsize 13/5/0} & {\scriptsize 11/7/0} & {\scriptsize 6/12/0} & {\scriptsize 12/6/0} & {\scriptsize 162/54/0} \\
\multicolumn{2}{l}{best(dice) $-$ best(loss)} & $+0.143$ & $+0.135$ & $+0.070$ & $+0.037$ & $+0.020$ & $+0.022$ & $+0.009$ & $+0.011$ & $-0.006$ & $+0.001$ & $+0.001$ & $-0.001$ & $+0.037 \pm 0.004$ \\
\multicolumn{2}{l}{\quad\scriptsize W/L/T} & {\scriptsize 18/0/0} & {\scriptsize 17/0/1} & {\scriptsize 15/1/2} & {\scriptsize 13/2/3} & {\scriptsize 14/1/3} & {\scriptsize 13/2/3} & {\scriptsize 10/5/3} & {\scriptsize 12/2/4} & {\scriptsize 8/10/0} & {\scriptsize 8/7/3} & {\scriptsize 7/8/3} & {\scriptsize 10/8/0} & {\scriptsize 145/46/25} \\
\multicolumn{2}{l}{best(dice) $-$ best(CE)} & $+0.012$ & $+0.006$ & $+0.008$ & $+0.008$ & $+0.014$ & $+0.010$ & $+0.008$ & $+0.011$ & $+0.005$ & $+0.002$ & $+0.004$ & $+0.003$ & $+0.008 \pm 0.001$ \\
\multicolumn{2}{l}{\quad\scriptsize W/L/T} & {\scriptsize 9/4/5} & {\scriptsize 8/5/5} & {\scriptsize 7/4/7} & {\scriptsize 8/3/7} & {\scriptsize 10/1/7} & {\scriptsize 11/4/3} & {\scriptsize 8/2/8} & {\scriptsize 11/1/6} & {\scriptsize 9/7/2} & {\scriptsize 7/7/4} & {\scriptsize 6/4/8} & {\scriptsize 7/4/7} & {\scriptsize 101/46/69} \\
\hline
\end{tabular}
\caption{
\textbf{Effect of using DSC as model selection rules on the label efficiency experiment.} Each cell reports the difference in test performances obtained by comparing
each model selected by the best Dice Score on the full validation pool, against taking the last checkpoint (selected by early stopping with $\text{patience} = 20$), the best validation loss model - i.e., the same approach used in \epsSeg-, or the model with the best cross-entropy term.
}
\label{tab:selection-rules}
\end{table*}

\miniheadline*{Selection rule of the final models}
All \epsSeg-based models of Table~\ref{tab:main-epssegpp-results}, \epsSeg included, are selected with the same rule, the top-$5$ ensemble.
Supp.\ Tab.~\ref{tab:strategy} reports the alternatives.
At a single operating point no rule is consistently best, and the differences reach up to $0.037$ mean Dice.
We use the ensemble because of its behaviour over many validation draws, where it removes the worst outcomes (Supp.\ Sec.~\ref{sec:selection_details}), not because it wins at every operating point.
\begin{table}[t]
\centering
\small
\setlength{\tabcolsep}{5pt}
\begin{tabular}{l!{\color{gray!50}\vrule}ccc}
\hline
Model & last & best on $V$ & top-$5$ ens. \\
\hline
\multicolumn{4}{l}{\textit{$10K$ annotated voxels}} \\
2D \epsSeg       & $0.746$ & $\mathbf{0.783}$ & $0.778$ \\
3D \epsSeg       & $0.865$ & $0.866$ & $\mathbf{0.883}$ \\
2D \epsSegpp     & $0.786$ & $0.805$ & $\mathbf{0.824}$ \\
3D \epsSegpp     & $\mathbf{0.890}$ & $0.882$ & $0.874$ \\
\multicolumn{4}{l}{\textit{extended annotation (matching baseline annotations)}} \\
2D \epsSeg       & $0.810$ & $0.797$ & $\mathbf{0.822}$ \\
3D \epsSeg       & $\mathbf{0.889}$ & $0.874$ & $0.879$ \\
2D \epsSegpp     & $0.813$ & $\mathbf{0.838}$ & $0.837$ \\
3D \epsSegpp     & $0.906$ & $\mathbf{0.919}$ & $0.905$ \\
\hline
\end{tabular}
\caption{\textbf{Checkpoint-selection rules for the \epsSeg-based models of Table~\ref{tab:main-epssegpp-results}.}
Mean Dice on the common region of the test section for the last checkpoint, the single best checkpoint on $V$, and the top-$5$ ensemble on $V$; best per row in bold.
At a single operating point no rule wins consistently: the ensemble is best in three of the eight rows, the last checkpoint in two, and the single best checkpoint in three.}
\label{tab:strategy}
\end{table}

\section{Checkpoint-Selection Strategies}
\label{sec:selection_details}

\miniheadline*{Setup}
The study uses our 2D \epsSegpp PoE model and the training approach of Section~\ref{sec:selection}: every run stores its checkpoints, and selection is applied \textit{post hoc}.
To make a validation-free training loop possible while keeping the training behaviour comparable, we replace the ReduceLROnPlateau scheduler of the label-efficiency experiment with a fixed exponential schedule that approximates its median run (Supp.\ Sec.~\ref{sec:training_details}).
We compare (i) the last checkpoint, which uses no validation set, (ii) the single checkpoint with the best Dice score on a validation set $V$, and (iii) the top-$5$ ensemble of Eq.~\eqref{eq:ensemble} on the same $V$.

\miniheadline*{Full pool versus smaller validation sets}
The advantage of a smaller validation set over the full pool is confined to training budgets between $5{,}000$ and $40{,}000$ annotations.
Below that range, a validation set of a quarter of the training budget holds only a few hundred annotations, and the whole pool remains the better choice.

\begin{figure*}[t]
    \centering
    \includegraphics[width=\textwidth]{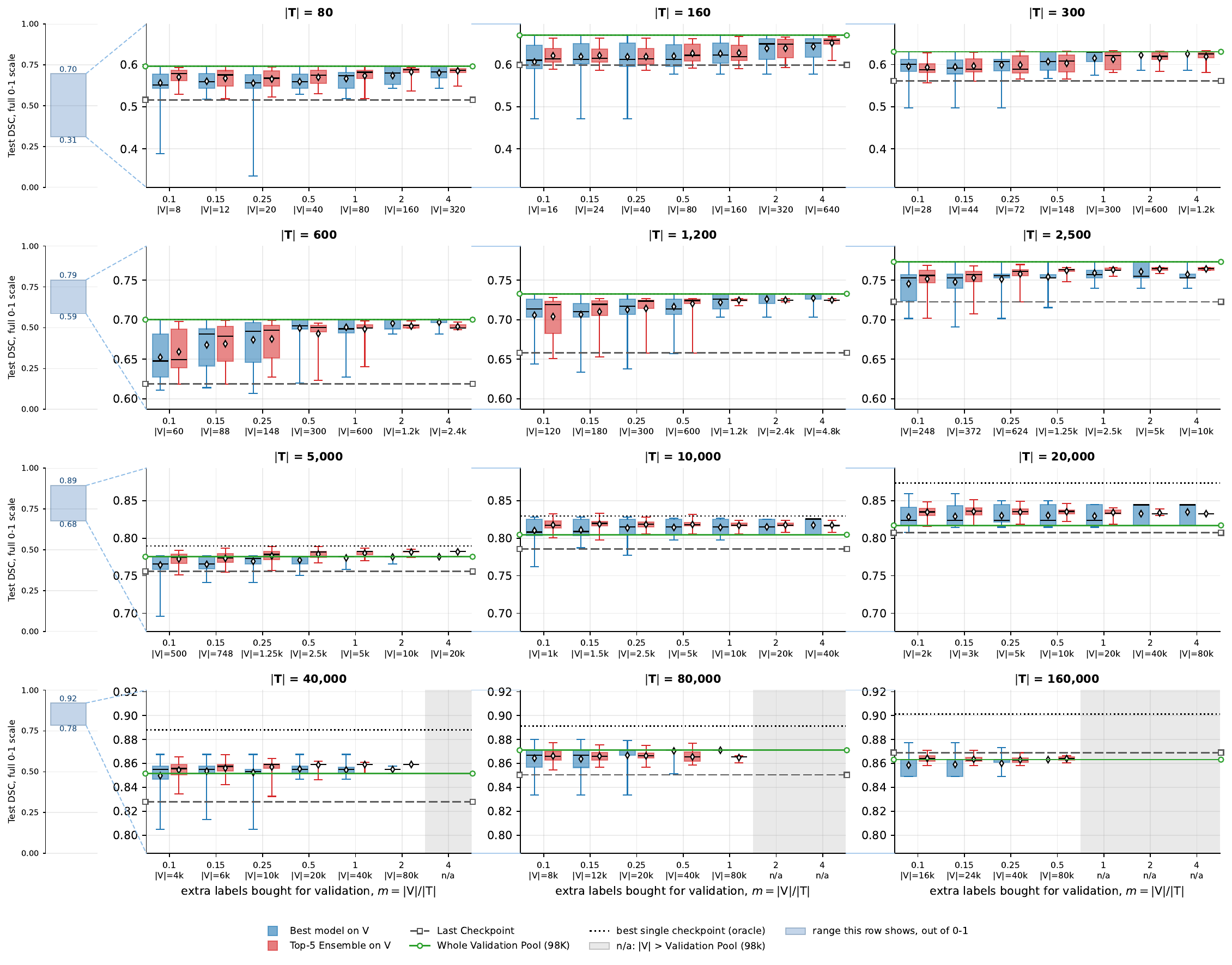}
    \caption{\textbf{Checkpoint selection at realistic validation-set sizes} (2D \epsSegpp, PoE).
    Each panel is one training budget $\lvert T\rvert$; within a panel, columns are nested validation sets of $m = \lvert V\rvert/\lvert T\rvert \in \{0.1, 0.15, 0.25, 0.5, 1, 2, 4\}$ times the training budget, each drawn $50$ times from the validation pool, so that every draw stands for one simulated user.
    Boxes show the test Dice (common region of section $z=626$) of the checkpoint chosen on each draw by the single best Dice on $V$ (blue) and by our top-$5$ ensemble on $V$ (red).
    Lines: best checkpoint on the whole validation pool of $98{,}000$ annotations (green), last checkpoint (dashed), and best saved checkpoint on the test section, \ie, the oracle (dotted).
    Grey columns would need more validation annotations than the pool holds.
    The small axis left of each row locates the row's zoomed range on the full $0$--$1$ scale.}
    \label{fig:selection_strategies}
\end{figure*}

\miniheadline*{Ensemble versus single best checkpoint}
Over the $84$ combinations of training and validation budget, each drawn $50$ times, the ensemble is ahead on the mean in $65$, and in $27$ of the $28$ combinations between $5{,}000$ and $40{,}000$ training annotations, but only by a median of $+0.005$ mean Dice.
On the worst draw it is ahead in $72$ of the $84$ combinations, and in all of them whenever the validation set is at most a quarter of the training set, by a median of $+0.021$ and by as much as $+0.19$ mean Dice.
The ordering is clearest where the validation set is smallest: at $\lvert V\rvert = 0.1\,\lvert T\rvert$ the ensemble is ahead on the worst draw at all twelve training budgets, by $+0.141$ and $+0.118$ mean Dice at the two smallest.
Supp.\ Fig.~\ref{fig:selection_strategies} shows the full grid of budgets and validation sizes.

\section{Per-Level Experts}
\label{sec:per_level}
\begin{figure*}[t]
    \centering
    \begin{minipage}{0.49\textwidth}
        \centering
        \includegraphics[width=\linewidth]{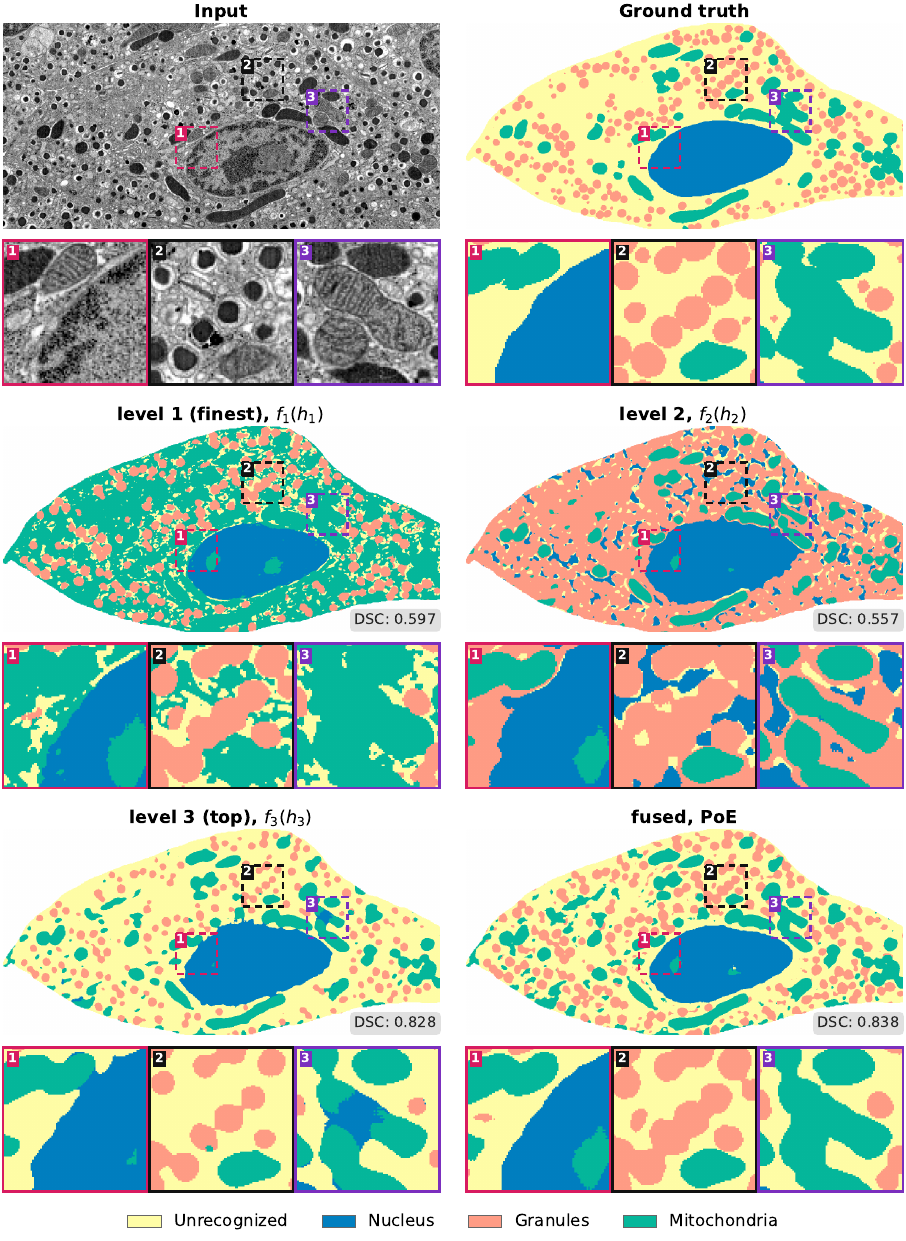}
    \end{minipage}\hfill
    \begin{minipage}{0.49\textwidth}
        \centering
        \includegraphics[width=\linewidth]{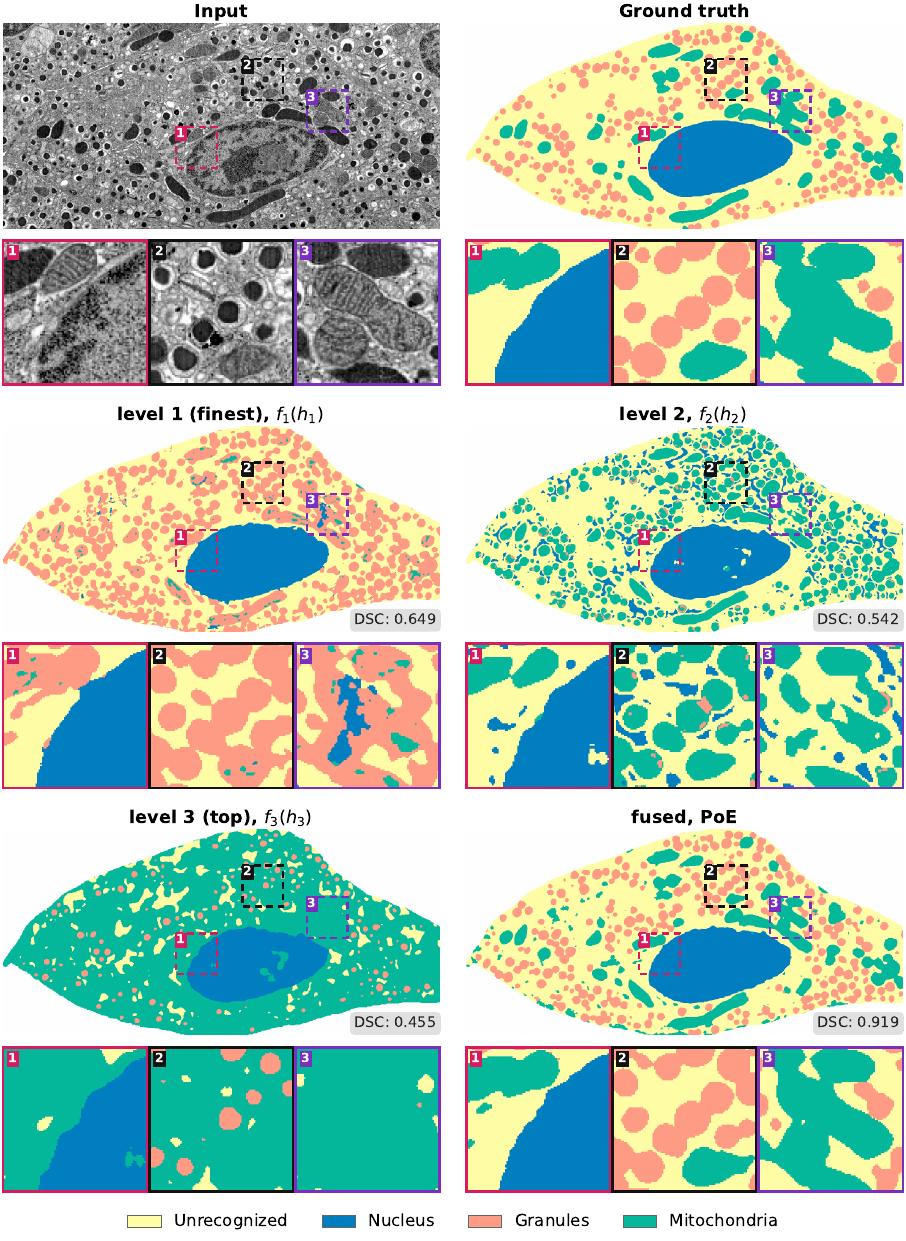}
    \end{minipage}
    \caption{\textbf{Per-level predictions of \epsSegpp} in (a) 2D and (b) 3D, trained on the extended annotation set and shown for one member of the top-$5$ checkpoint ensemble on the test section ($z=626$, common region).
    For each model: input and ground truth, the prediction of each level's expert alone (level 1 is the finest), and the fused product-of-experts prediction; badges give the mean Dice of each panel.
    Since the experts are trained only through their fused prediction, a single level's prediction is not uniquely determined and is shown as an illustration, not as a model of its own.}
    \label{fig:per_level}
\end{figure*}

\begin{table*}[t]
\centering
\small
\setlength{\tabcolsep}{5pt}
\begin{tabular}{l!{\color{gray!50}\vrule}cccc!{\color{gray!50}\vrule}c}
\hline
Level & U & N & G & M & Avg. Dice \\
\hline
\multicolumn{6}{l}{\textit{2D \epsSegpp, extended annotation}} \\
level 1      & $0.361$ & $0.915$ & $0.794$ & $0.354$ & $0.606$ \\
level 2      & $0.489$ & $0.764$ & $0.571$ & $0.807$ & $0.658$ \\
level 3      & $0.867$ & $0.933$ & $0.727$ & $0.774$ & $0.825$ \\
\rowcolor{gray!15}
fused (PoE)  & $0.861$ & $0.948$ & $0.779$ & $0.752$ & $0.835$ \\
\hline
\multicolumn{6}{l}{\textit{3D \epsSegpp, extended annotation}} \\
level 1      & $0.857$ & $0.954$ & $0.599$ & $0.131$ & $0.635$ \\
level 2      & $0.789$ & $0.635$ & $0.084$ & $0.456$ & $0.491$ \\
level 3      & $0.322$ & $0.929$ & $0.302$ & $0.287$ & $0.460$ \\
\rowcolor{gray!15}
fused (PoE)  & $0.911$ & $0.970$ & $0.896$ & $0.808$ & $0.896$ \\
\hline
\end{tabular}
\caption{\textbf{Per-level predictions (illustration).} Per-class Dice of each level's expert alone and of their product, for 2D and 3D \epsSegpp trained on the extended annotation set ($\lvert T\rvert = 25{,}822$, $\lvert V\rvert = 16{,}069$) and scored on the common region of the test section ($z=626$). Each entry is the mean over the five members of the top-$5$ checkpoint ensemble, scored one checkpoint at a time; Avg.\ Dice is the unweighted mean over the four classes. Shaded rows are the fused prediction the model outputs. Only the sum of the levels' logits enters the classification loss, so the single-level rows are not uniquely determined by training; they illustrate how this model distributes its evidence and are not the performance of independent classifiers.}
\label{tab:per_level}
\end{table*}

Supp.\ Fig.~\ref{fig:per_level} and Supp.\ Tab.~\ref{tab:per_level} show what each expert predicts on its own, for the 2D and 3D models of Table~\ref{tab:main-epssegpp-results} trained on the extended annotation.
These predictions are an illustration of one trained model, not a measurement of what each level has learned.
The classification loss sees only the sum of the levels' logits, so a class-wise offset added to one level and subtracted from another leaves the loss unchanged.
Each expert's logits also condition its own level through FiLM, which constrains them further, but a single level's prediction remains only weakly determined by training and can differ between runs that reach the same fused prediction.
With this caveat, the fused prediction is better in mean Dice than every single level in both models; in 3D, no single level exceeds $0.635$, yet the fused prediction reaches $0.896$.

\section{Qualitative Results}
\label{sec:qualitative_details}
\begin{figure*}[p]
    \centering
    \includegraphics[width=0.86\textwidth]{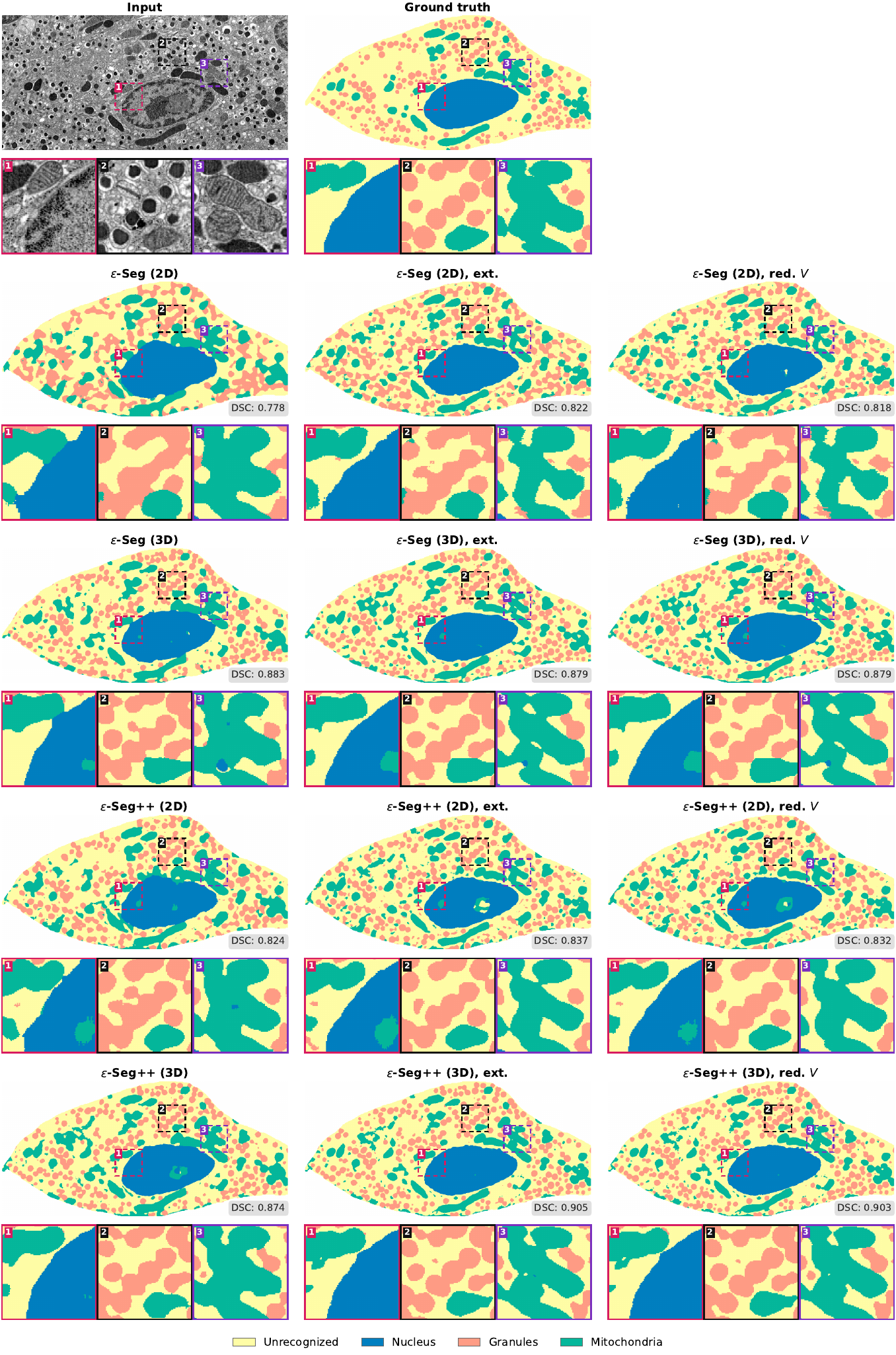}
    \caption{\textbf{Qualitative results of all \epsSeg-based models} on the test section ($z=626$).
    Rows: 2D and 3D \epsSeg and \epsSegpp.
    Columns: the three annotation settings of Table~\ref{tab:main-epssegpp-results}, \ie, patch centres only, the same annotation as the baselines (ext.), and the same training annotation validated on patch centres only (red.\ $V$).
    Insets as in Figure~\ref{fig:seg_models}; badges give the mean Dice on the common region.}
    \label{fig:supp_qualitative}
\end{figure*}

\end{document}